\documentclass{article}

\PassOptionsToPackage{numbers, compress}{natbib}

\usepackage[preprint]{neurips_2024}

\usepackage[utf8]{inputenc} 
\usepackage[T1]{fontenc}    
\usepackage{url}            
\usepackage{booktabs}       
\usepackage{amsfonts}       
\usepackage{nicefrac}       
\usepackage{microtype}      

\usepackage[pdftex,dvipsnames]{xcolor}  
\usepackage{xargs}
\usepackage[colorinlistoftodos,prependcaption,textsize=tiny]{todonotes}
\newcommandx{\xy}[2][1=]{\todo[linecolor=red,backgroundcolor=red!25,bordercolor=red,#1]{#2}}
\newcommandx{\change}[2][1=]{\todo[linecolor=blue,backgroundcolor=blue!25,bordercolor=blue,#1]{#2}}
\newcommandx{\info}[2][1=]{\todo[linecolor=OliveGreen,backgroundcolor=OliveGreen!25,bordercolor=OliveGreen,#1]{#2}}
\newcommandx{\improvement}[2][1=]{\todo[linecolor=Plum,backgroundcolor=Plum!25,bordercolor=Plum,#1]{#2}}
\newcommandx{\thiswillnotshow}[2][1=]{\todo[disable,#1]{#2}}
\usepackage{graphicx}
\usepackage{multirow} 
\usepackage{tcolorbox}
\usepackage{natbib}
\usepackage{amsmath}
\usepackage{amssymb}
\usepackage{multirow}
\usepackage{graphics}
\usepackage{threeparttable}
\usepackage{color}
\usepackage[normalem]{ulem}
\usepackage{float}
\usepackage{amsfonts}
\usepackage{bm}
\usepackage{bbm}
\usepackage{enumitem}
\usepackage{wrapfig}
\usepackage{natbib}
\usepackage{subcaption}
\usepackage{algorithm}
\usepackage{array}
\usepackage{makecell}
\usepackage{color, colortbl}
\usepackage{pifont}
\usepackage{booktabs}
\usepackage{mathtools}
\usepackage{siunitx}
\usepackage{caption} 
\usepackage{xcolor}
\usepackage{subcaption} 
\usepackage[nopar]{lipsum}
\usepackage{listings}
\usepackage[export]{adjustbox}
\usepackage{makecell}
\usepackage{wrapfig}
\usepackage{mathtools}
\usepackage{siunitx}
\usepackage[nopar]{lipsum}
\usepackage{listings}

\usepackage{enumitem}
\usepackage{algpseudocode}
\usepackage[font=small,labelfont=bf]{caption}
\usepackage{array}
\usepackage{multirow}
\usepackage{booktabs}
\usepackage{algorithm}
\usepackage{subcaption}
\usepackage[normalem]{ulem}
\usepackage{xparse}
\usepackage{pifont}
\usepackage{bm}
\usepackage{threeparttable}
\usepackage{etoolbox}

\usepackage{listings}
\usepackage{mwe} %
\usepackage{makecell}
\usepackage{color, colortbl}
\usepackage{tabularx}
\usepackage{pifont}%
\usepackage[accsupp]{axessibility}
\usepackage{listings}
\definecolor{myy}{RGB}{126,95,0}
\definecolor{mygray}{gray}{.9}
\definecolor{Gray}{gray}{0.9}
\definecolor{bblue}{RGB}{30,80,120}
\definecolor{mygray1}{gray}{.7}
\definecolor{ggray}{RGB}{127,127,127}
\definecolor{defaultcolor}{gray}{.9}
\definecolor{dark-gray}{gray}{0.20}
\newcommand{\reshl}[2]{
	\textbf{#1} \fontsize{7.5pt}{1em}\selectfont\color{mygreen}{$\uparrow$ \textbf{#2}}
}

\definecolor{mygreen}{HTML}{39b54a}
\newcommand{\pub}[1]{{\color{dark-gray}{\tiny{[{#1}]}}}}

\newcolumntype{x}[1]{>{\centering\arraybackslash}p{#1pt}}
\newcolumntype{y}[1]{>{\raggedright\arraybackslash}p{#1pt}}
\newcolumntype{z}[1]{>{\raggedleft\arraybackslash}p{#1pt}}

\definecolor{cvprblue}{rgb}{0.21,0.49,0.74}
\usepackage{color}

\definecolor{micoblue}{HTML}{155AE5}
\definecolor{micopurple}{HTML}{B060E2}
\definecolor{micogreen}{HTML}{8FE260}
\definecolor{micoyellow}{HTML}{DEE260}
\definecolor{micored}{HTML}{E26060}
\newcommand{\mico}{{\textbf{\textcolor{micored}{M}\textcolor{micoyellow}{i}\textcolor{micopurple}{C}\textcolor{micogreen}{o}}}\xspace}

\usepackage{enumitem}
\usepackage{algpseudocode}
\usepackage[font=small,labelfont=bf]{caption}
\usepackage{array}
\usepackage{multirow}
\usepackage{booktabs}
\usepackage{subcaption}
\usepackage[normalem]{ulem}
\usepackage{xparse}
\usepackage{pifont}
\usepackage{bm}
\usepackage{threeparttable}
\usepackage{etoolbox}
\usepackage{listings}
\usepackage{mwe} %
\usepackage{makecell}
\usepackage{color, colortbl}
\usepackage{tabularx}
\usepackage{pifont}%
\usepackage[accsupp]{axessibility}

\usepackage[scaled=0.85]{DejaVuSansMono}

\definecolor{citecolor}{HTML}{0071bc}

\usepackage{graphicx}
\usepackage{amsmath}
\usepackage{amssymb}
\usepackage{pgfplots}
\usepackage{multirow}
\pgfplotsset{compat=1.16}
\usepgfplotslibrary{groupplots}
\usetikzlibrary{patterns}

\newlength\savewidth\newcommand\shline{\noalign{\global\savewidth\arrayrulewidth
  \global\arrayrulewidth 1pt}\hline\noalign{\global\arrayrulewidth\savewidth}}

\newlength\thinwidth
\definecolor{Gray}{gray}{0.92}
\definecolor{DarkGray}{gray}{0.5}
\newcolumntype{H}{>{\setbox0=\hbox\bgroup}c<{\egroup}@{}}
\definecolor{LightCyan}{rgb}{0.88,1,1}
\definecolor{altRowColor}{gray}{0.92}
\definecolor{highlightRowColor}{rgb}{0.9, 0.9, 1}

\definecolor{GrayNumber}{gray}{0.5}
\definecolor{GrayXMark}{gray}{0.7}
\definecolor{TextDark}{rgb}{1,0.27,0}
\definecolor{ImageDark}{rgb}{0,0.3,0.8}
\definecolor{VideoDark}{rgb}{.5,.0,.5}
\definecolor{DepthDark}{rgb}{0,.5,0}
\definecolor{AudioDark}{rgb}{0.992156862745098, 0.7686274509803922, 0.0941176470588235}
\definecolor{ThermalDark}{rgb}{0.8823529411,0.63725490196,0.0156862745}
\definecolor{IMUDark}{rgb}{0.6235294117647059, 0.27058823529411763, 0.4627450980392157}
\definecolor{TimeDark}{HTML}{45DC61}  
\definecolor{GraphDark}{HTML}{00A6FC}
\definecolor{HyperDark}{HTML}{D1A44F}
\definecolor{TabularDark}{HTML}{FF7900}
\colorlet{Image}{ImageDark!20!white}
\colorlet{Video}{VideoDark!20!white}
\colorlet{Depth}{DepthDark!20!white}
\colorlet{Audio}{AudioDark!20!white}
\colorlet{ImageLight}{ImageDark!70!white}
\colorlet{VideoLight}{VideoDark!70!white}
\colorlet{DepthLight}{DepthDark!70!white}
\colorlet{AudioLight}{AudioDark!70!white}
\newcommand{\symbolHt}{1.5em}
\newcommand{\textChar}{%
  \begingroup\normalfont
  \includegraphics[height=\symbolHt]{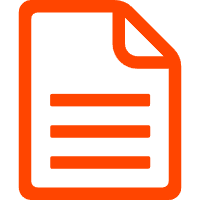}%
  \endgroup
}
\newcommand{\imageChar}{%
  \begingroup\normalfont
  \includegraphics[height=\symbolHt]{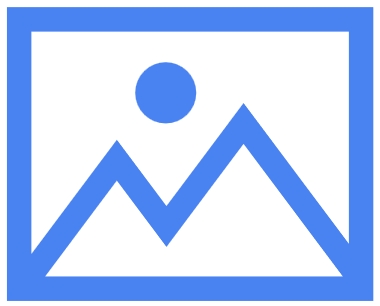}%
  \endgroup
}
\newcommand{\videoChar}{%
  \begingroup\normalfont
  \includegraphics[height=\symbolHt]{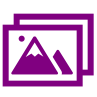}%
  \endgroup
}
\newcommand{\timeChar}{%
  \begingroup\normalfont
  \includegraphics[height=\symbolHt]{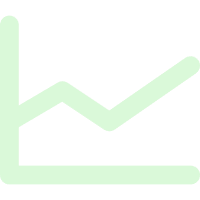}%
  \endgroup
}
\newcommand{\audioChar}{%
  \begingroup\normalfont
  \includegraphics[height=\symbolHt]{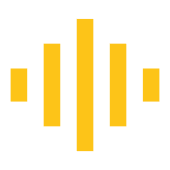}%
  \endgroup
}
\newcommand{\depthChar}{%
  \begingroup\normalfont
  \includegraphics[height=\symbolHt]{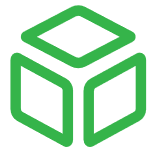}%
  \endgroup
}
\newcommand{\thermalChar}{%
  \begingroup\normalfont
  \includegraphics[height=\symbolHt]{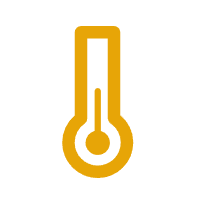}%
  \endgroup
}
\newcommand{\imuChar}{%
  \begingroup\normalfont
  \includegraphics[height=\symbolHt]{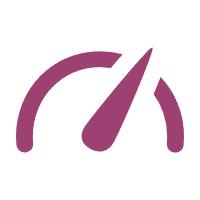}%
  \endgroup
}
\newcommand{\graphChar}{%
  \begingroup\normalfont
  \includegraphics[height=\symbolHt]{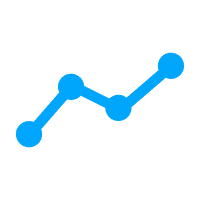}%
  \endgroup
}
\newcommand{\hyperChar}{%
  \begingroup\normalfont
  \includegraphics[height=\symbolHt]{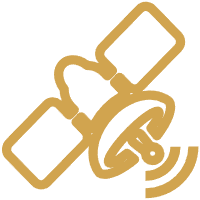}%
  \endgroup
}
\newcommand{\tabularChar}{%
  \begingroup\normalfont
  \includegraphics[height=\symbolHt]{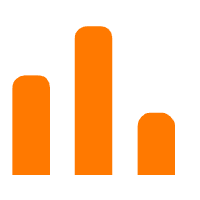}%
  \endgroup
}
\newcolumntype{i}{>{\columncolor{Image}}c}
\newcolumntype{v}{>{\columncolor{Video}}c}
\newcolumntype{d}{>{\columncolor{Depth}}c}
\newcolumntype{a}{>{\columncolor{Audio}}c}
\newcolumntype{I}{>{\columncolor{ImageLight}}c}
\newcolumntype{D}{>{\columncolor{DepthLight}}c}
\newcolumntype{A}{>{\columncolor{AudioLight}}c}
\newcolumntype{E}{>{\columncolor{highlightRowColor}}c}

\usepackage[pagebackref,breaklinks,colorlinks,citecolor=cvprblue]{hyperref}
\title{Explore the Limits of Omni-modal Pretraining at Scale}

\author{
~~~~ {Yiyuan Zhang}$^{1,4*} $ 
~~~  {Handong Li}$^{2,3}$\thanks{Equal Contribution}
~~~~ {Jing Liu}$^{2,3}$\thanks{Corresponding Author}
~~~ {Xiangyu Yue}$^{1}$ \\
\textsuperscript{1} Multimedia Lab, The Chinese University of Hong Kong \\
\textsuperscript{2} School of Artificial Intelligence, University of Chinese Academy of Sciences \\
\textsuperscript{3} Institute of Automation, Chinese Academy of Science 
\textsuperscript{4} Shanghai AI Laboratory\\
}

\begin{document}

\maketitle

\vspace{-3mm}
\begin{abstract}
    We propose to build omni-modal intelligence, which is capable of understanding any modality and learning universal representations. In specific, we propose a scalable pretraining paradigm, named \textbf{M}ult\textbf{i}modal \textbf{C}\textbf{o}ntext (\mico), which can scale up the numbers of modalities and amount of data, together with the model parameters, in the pretraining process. With \mico, the pretrained models show significant emergent abilities in multimodal learning, which are evaluated on the following tasks: i) single-modality perception benchmarks of 10 different modalities, ii) 25 cross-modality understanding tasks of retrieval, question-answering, captioning, and iii) 18 multimodal large language model benchmarks. Our models establish \textbf{37} new records for state-of-the-art performance.   We hope that our research could contribute to the development of omni-modal intelligence. \href{https://github.com/invictus717/MiCo}{\textcolor{red}{\textbf{Code and Models}}}
\end{abstract}

\section{Introduction}~\label{sec:intro}
In the development of artificial intelligence, scalable pre-training has emerged as a promising pathway towards general intelligence~\cite{radford2019language,OpenAI2023GPT4TR,brown2020language,radford2021learning,bubeck2023sparks}. 
\begin{figure*}[ht]
\centering
\vspace{-3.5mm}
\includegraphics[width=0.98\linewidth]{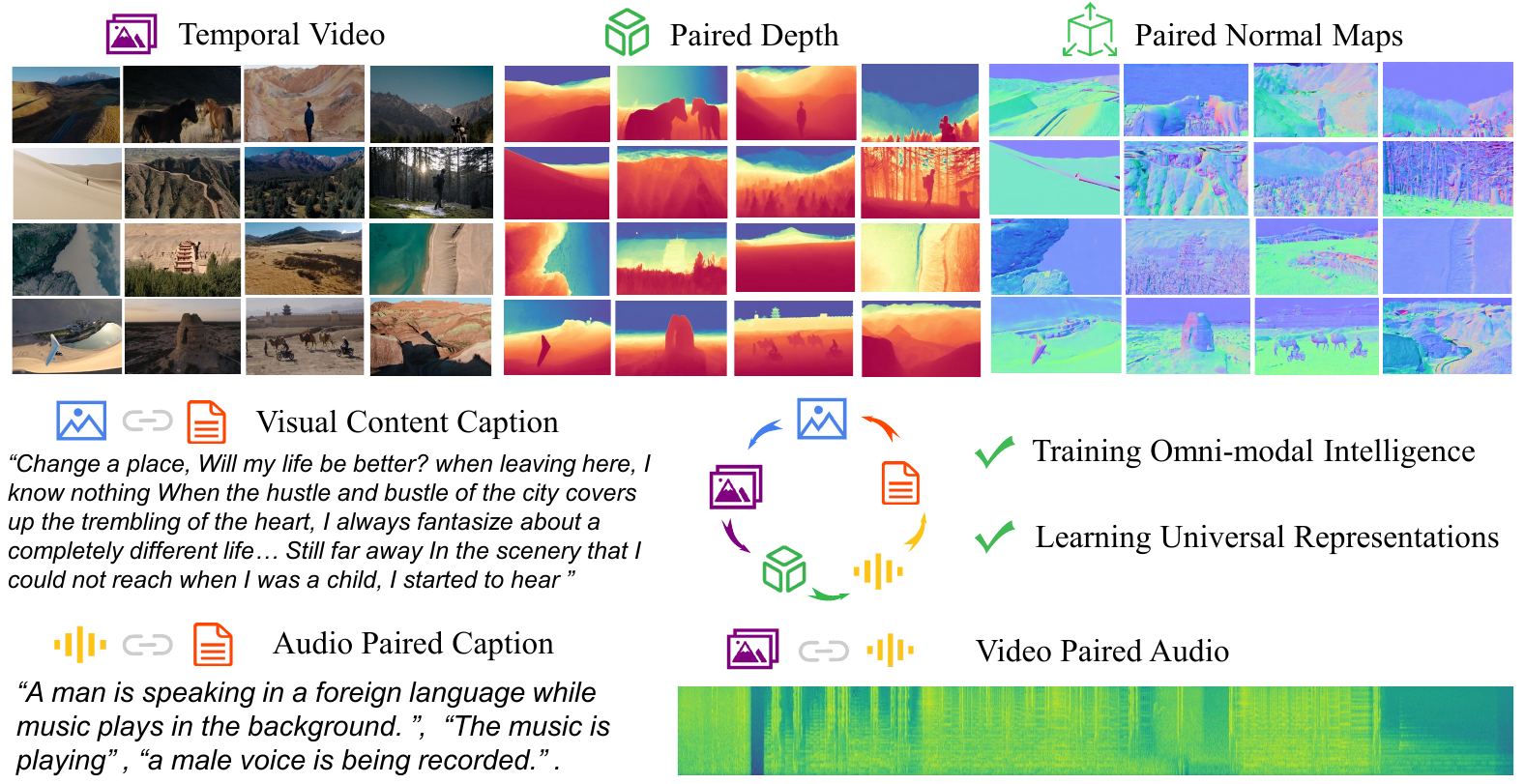}
\vspace{-1mm}
\caption{\textbf{Omni-modal Pretraining}. We propose collecting large-scale omni-modal paired data, including text, image, video, depth, and normal maps, to learn universal representations.}
    \label{fig:data}
\vspace{-3mm}
\end{figure*}
  Additionally, pre-training has been established as an effective approach for learning more general and transferable representations across various modalities. For example, CLIP~\cite{radford2021learning} constructs million-scale text-image pairs for cross-modal contrastive learning, making it one of the most impactful foundation models in the community~\cite{rombach2022high,poole2022dreamfusion}. Researchers have further extended the capabilities of CLIP~\cite{radford2021learning} to more data modalities, \textit{e.g.} audio~\cite{guzhov2022audioclip}, point clouds~\cite{xue2023ulip}, and more comprehensive tasks, \textit{e.g.} reasoning about images/ videos with large language models (LLMs)~\cite{liu2024visual,2023videochat}. The main contributions of CLIP~\cite{radford2021learning} are two-fold: collecting web-scale text-image data and proposing a scalable vision-language pretraining paradigm. As more modalities \textit{e.g.} audio, video, and 3D content, are getting widely used in this multimodal era~\cite{han2023onellm,2023videochat,ding2023unireplknet,girdhar2023imagebind,rombach2022high,poole2022dreamfusion}, such developments present additional challenges, including multimodal misalignment, misinterpretation, and bias amplification, in achieving coherent multimodal understanding with LLMs.

\begin{figure*}[ht]
\centering
\vspace{-3mm}
\includegraphics[width=0.93\linewidth]{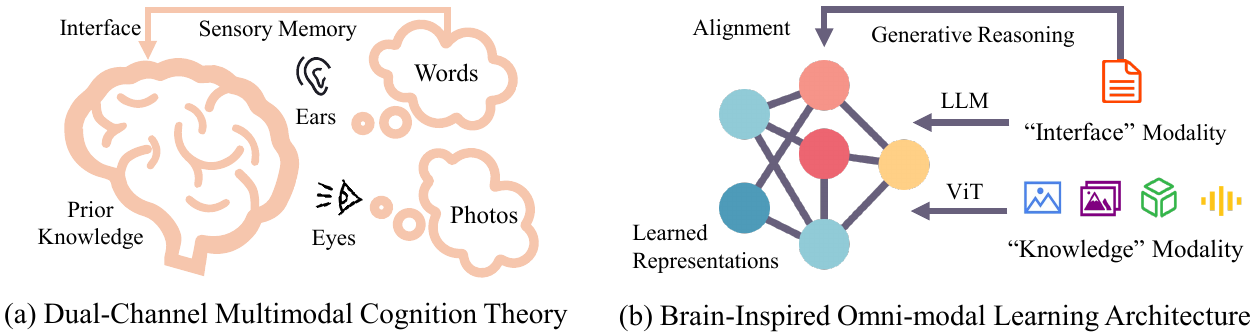}
\vspace{-1mm}
\caption{\textbf{Multimedia Cognition Process in Brain Inspires our Design}. We split diverse modalities into two types and employ individual neural networks to learn representations from each type respectively.}
\vspace{-1.1mm}
\label{fig:cognition}
\end{figure*}

In this paper, we aim to enhance the comprehensive abilities of CLIP in visual understanding and further bolster its multimodal capacities across audio, video, 3D content, and more, as illustrated in Figure~\ref{fig:data}. This is significantly challenging. Therefore, we shift our focus from training a general multimodal model to understanding how the human brain performs coherent multimodal cognition. As outlined in Richard Mayer's Cognitive Theory of Multimedia Learning~\cite{mayer2002multimedia}, our brain processes multimedia signals through two distinct channels—auditory and visual—in sensory memory, as depicted in Figure~\ref{fig:cognition}. The sensory memory integrates these signals with prior knowledge through words, transforming new multimedia information into long-term memory. Notably, \textbf{1}) multimedia signals in the brain share channels, and \textbf{2}) words function as the reasoning interface in our brain.

Inspired by these insights, we categorize diverse modalities into two types: ``knowledge modality'' and ``interface modality''. \textit{Knowledge modalities}, primarily derived from raw sensors, contribute knowledge in diverse formats. For example, images and depth maps offer visual knowledge, while audio and video provide auditory and spatiotemporal knowledge. The language modality, developed by humans, is inherently more abstract and naturally functions as the \textit{interface modality}, facilitating learning, reasoning, and the coordination of knowledge.
To this end, we design an omni-modal learning architecture, illustrated in Figure~\ref{fig:cognition} (b), with two distinct branches: one for knowledge modalities and one for the interface modality, \textit{i.e.} natural language. The knowledge and interface modalities are aligned through a novel generative reasoning method, as detailed in \S~\ref{sec:method:modal}.

\begin{figure*}[t]
\centering
\includegraphics[width=0.98\linewidth]{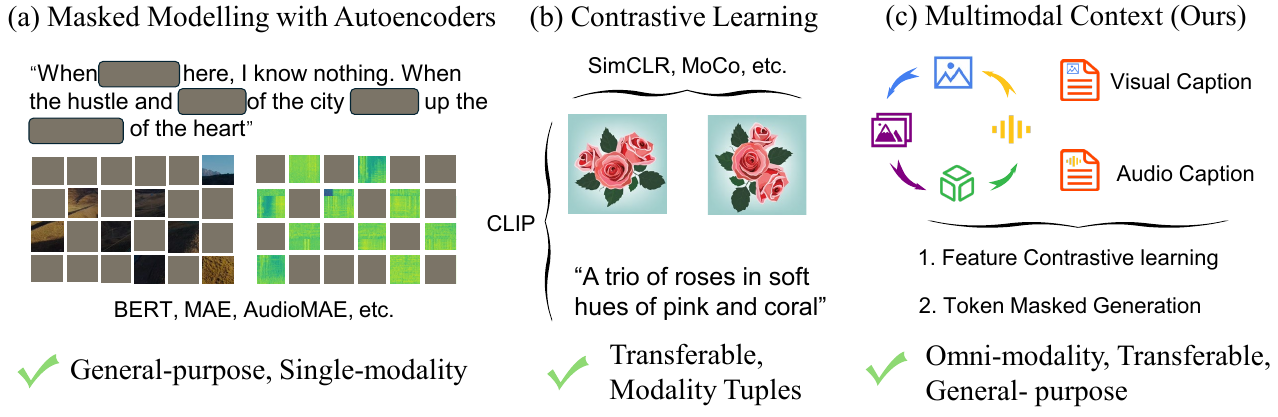}
\caption{\textbf{Evolution of Pretraining Paradigms}. Masked modeling ~\cite{he2022masked,huang2022masked,devlin2018bert} has shown great success in single-modality general-purpose understanding. Contrastive learning~\cite{he2020momentum,radford2021learning,chen2020simple} distinguishes transferable features with modality tuples. We aim to achieve general-purpose omni-modal understanding and learn transferable, universal representations. }
\vspace{-6mm}
\label{fig:evolution}
\end{figure*}

In addition to the architecture design, the next challenge is how to further enhance the benefits of integrating multiple data modalities. The key to learning token sequences is the context relationship~\cite{vaswani2017attention}, which assigns a unique vector to each input position in a sequence. 
This approach improves sequence modeling by capturing the sequential relationship among tokens.
Moreover, since different modalities (\textit{e.g.}, text, image, audio) offer complementary information, integrating these sources fosters a more comprehensive understanding of the data. Modeling token sequences from different modalities under the same context can help the model understand modality characteristics and joint semantics.

Therefore, we propose a Multimodal Context (\mico) framework. We first map different modalities into a joint embedding space by sharing backbone networks. Then we build contextual relationships by sharing the same position embeddings and utilizing additional context embeddings to enhance coherent multimodal understanding, as shown in Figure~\ref{fig:data}. Subsequently, we employ omnimodal contrastive learning, omnimodal feature matching, and omnimodal caption generation processes for pretraining (detailed in \S~\ref{sec:method:obj}). Moreover, \mico can incorporate existing text-image, text-audio, and text-video datasets for joint multimodal context learning (\S~\ref{sec:method:modal}), which leads to better omni-modal learning capacity, further modality extensibility, and easier scalability of multimodal data. Meanwhile, we are the first to explore \textbf{multimodal scaling laws} in pretraining modalities, model parameters, and data scales (detailed in Figure~\ref{fig:scaling_law}).

As shown in Figure~\ref{fig:evolution}, we compare \mico with existing pretraining approaches. With omnimodal contrastive learning, omnimodal feature matching, and omnimodal caption generation processes, \mico successfully integrates the advantages of both masked modeling and contrastive learning. In other words, \mico represents the next-generation evolution of masked modeling~\cite{he2022masked,huang2022masked,devlin2018bert} and contrastive learning methods~\cite{he2020momentum,radford2021learning,chen2020simple} for the multimodal era, offering significant benefits in omni-modal learning, strong transferability, and general-purpose representations. To thoroughly evaluate the effectiveness of \mico, we conduct extensive experiments on universal single-modality perception benchmarks, cross-modal retrieval, captioning, and question-answer (QA) benchmarks, as well as zero-shot QA benchmarks for multimodal large language models. \mico achieves impressive results across these benchmarks, establishing more than \textbf{37} new state-of-the-art (SOTA) performances and showing remarkable improvements of over \textbf{20}\% on some benchmarks. These results compellingly illustrate that \mico is a promising next-generation pretraining paradigm for the multimodal era.

\vspace{-0.2cm}
\section{Related Work}~\label{sec:related}
\textbf{Vision-Language Pretraining.} 
MCAN~\cite{yu2019deep} first aligns vision and language features by stacking deep cross-attention blocks. Then more works~\cite{wang2021vlmo,wang2021simvlm,wang2022unifying,wang2022image} scale their models and improve the vision-language fusion process to build better alignment. VL-BERT~\cite{su2019vl} introduced the Masked Language Model (MLM) paradigm, focusing on generic tasks across both vision and language modalities. Then Oscar~\cite{li2020oscar} proposed to enrich the representation of object semantics by integrating visual and textual content. Subsequent frameworks have further refined and extended these capabilities. Notably, VinVL~\cite{zhang2021vinvl}, SimVLM~\cite{wang2021simvlm}, VLMO~\cite{wang2021vlmo}, ALBEF~\cite{li2021align}, and Florence~\cite{yuan2021florence} have explored and demonstrated the advantages of joint representations that ensure semantic consistency across the visual and natural language. Additionally, the versatility of multimodal models extends into specialized applications such as few-shot learning~\cite{alayrac2022flamingo}, and sequence-to-sequence~\cite{wang2022unifying,yu2022coca}. BEiT-v3~\cite{wang2022image} treats images as a ``foreign language'', employing a cross-modal mask-and-reconstruction process with partially shared parameters.

\textbf{More-Modality Pretraining}. MMV~\cite{alayrac2020self} pioneered multimodal pretraining using text, video, and audio pairs. They proposed multimodal contrastive learning for alignment.
Then VATT~\cite{akbari2021vatt} further developed pretraining multiple modalities with transformers. After CLIP~\cite{radford2021learning}, more works~\cite{zhang2023meta,girdhar2023imagebind,guzhov2022audioclip,xue2023ulip,xu2021videoclip,wang2024internvideo2} propose to adapt pretrained CLIP models to more modalities including point cloud, depth, audio, video, \textit{etc}. Another direction is to exploit multimodal complementary benefits and construct better and more modality pairs for pertaining foundation models such as VAST~\cite{chen2023vast} and VALOR~\cite{chen2023valor}, which improve the abilities for multimodal understanding.

Despite significant advancements in multimodal learning, several key challenges impede the development of comprehensive omni-modal intelligence: \textbf{1}) \textbf{Focus on Vision-Language Modalities}: Current methods~\cite{wang2022image,wang2021vlmo,li2020oscar,wang2021simvlm} predominantly cater to vision and language tasks. The inflexibility of these works limits the extension with more modalities such as video, depth, normals, and audio.
\textbf{2}) \textbf{Architectural Constraints}: The development of architectures capable of handling a broader array of modalities is still in its nascent stages. \textit{Crafting scalable and efficient multimodal learning architectures presents a significant challenge}.
\textbf{3}) \textbf{Data Availability}: There is a notable scarcity of publicly accessible multimodal datasets that include paired data (such as video, depth, audio, and captions). 
\textbf{4}) \textbf{Utilizing Multimodal Benefits}: Although leveraging the synergistic benefits of multiple modalities is crucial for achieving omni-modal intelligence~\cite{fei2022towards}, \textit{understanding and optimizing the interaction between highly disparate modalities remains a complex and largely unexplored area}.


\section{Multimodal Context}~\label{sec:method}
\vspace{-5mm}
\subsection{Large-Scale Data Collection}~\label{sec:method:data}
We use the HD-VILA~\cite{xue2022hdvila} dataset, which contains 371.5K hours of 720p (\(1280\times720\)) videos. We remove 
video clips that are shorter than 5s or longer than 30s. Then, we collect a dataset containing 1.7M paired video clips ($\sim$510M frames), audio, and subtitles \(\{(\boldsymbol{x}_V, \boldsymbol{x}_T^V, \boldsymbol{x}_A)\}\). Then we enrich the dataset by adding captions to video frames (images), and audio with pretrained captioners~\cite{chen2023vast}, getting \((\boldsymbol{x}_I, \boldsymbol{x}_T^I)\) and \((\boldsymbol{x}_A, \boldsymbol{x}_T^A)\). Finally, we use pretrained monocular depth estimation models~\cite{fu2024geowizard,eftekhar2021omnidata}\footnote{ Geowizard~\cite{fu2024geowizard} delivers significantly better annotations obviously, while DPT~\cite{eftekhar2021omnidata} predicts much faster (about 3\~4.7\(\times\) faster). We use the Geowizard to annotate the high-quality data about 2M.} to generate depth and normal maps, getting \((\boldsymbol{x}_I, \boldsymbol{x}_D, \boldsymbol{x}_N)\). Thus, we collect million-scale multimodal paired data \(\{(\boldsymbol{x}_I,\boldsymbol{x}_D, \boldsymbol{x}_N, \boldsymbol{x}_T^I), (\boldsymbol{x}_A,\boldsymbol{x}_T^A), (\boldsymbol{x}_V,\boldsymbol{x}_T^V)\}\), where \(\boldsymbol{x}_T, \boldsymbol{x}_I, \boldsymbol{x}_A, \boldsymbol{x}_V, \text{and } \boldsymbol{x}_D\) denote the modality-specific samples of text captions, image, audio, video clips, depth, and normal maps. We split our dataset into several subsets including 1M, 10M, 110M, and 334M multimodal data pairs, and we provide detailed illustrations in Appendix~\ref{sec:app:data}.

\subsection{Architecture Design for Omni-modal 
 Learning}~\label{sec:method:arch}
We first investigate several variants of encoder architectures with four data modalities. 
With our collected data, we pretrain architectures for 300K steps by the same contrastive~\cite{radford2021learning} and masked-generation loss functions~\cite{wang2022image} (details in Appendix~\ref{sec:app:train}). We take the captioning and retrieval tasks on image, audio, and video modalities as the main evaluation benchmark for designing architectures.

\textbf{Architectural Designs}. We construct the vanilla architecture from CLIP~\cite{radford2021learning}. A text encoder of Transformer~\cite{vaswani2017attention} takes text inputs and outputs text embeddings \(\boldsymbol{z}_T\), and an image encoder of Vision Transformer~\cite{dosovitskiy2020image} takes image input $x_I\in\mathbb{R}^{3\times H \times W}$ and outputs image embeddings \(\boldsymbol{z}_I\), respectively. 

\begin{figure}[ht]
    \centering
    \vspace{-4mm}
    \includegraphics[width=0.98\linewidth]{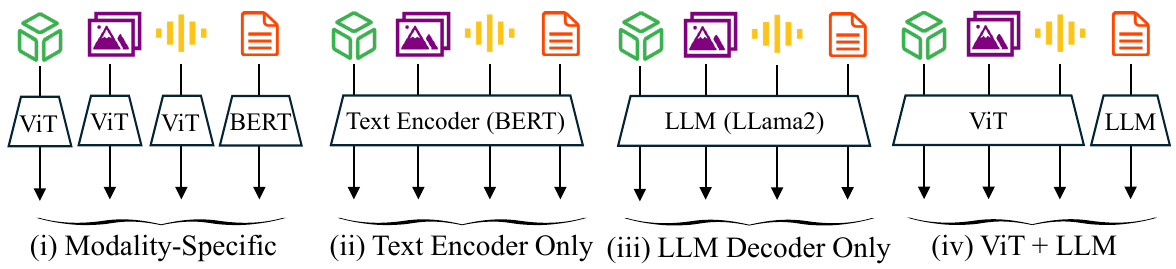}
    \vspace{-1mm}
    \caption{Options of Architecture Design for Omni-Modal Pretraining.}
    \vspace{-3mm}
    \label{fig:arch}
\end{figure}

As shown in Figure~\ref{fig:arch}, we propose 4 architectures for omni-modal learning: \textbf{i}) Modality-specific encoders for each modality, employing individual transformers to extract multimodal embeddings, then fuse them as BEiT-3~\cite{wang2022image}. \textbf{ii}) BERT (text encoder) as a unified multimodal encoder to extract multimodal embeddings and generates texts. \textbf{iii}) LLM (text decoder) as a unified multimodal encoder and text generator. \textbf{iv}) A ViT as a unified multimodal encoder besides text, and an LLM deals with text embeddings and generation.

\begin{table*}[ht]
\centering
\vspace{-2mm}
\caption{\textbf{Architecture Design of Omni-Modal Learning Paradigm}. The default backbone and LLM are ViT-g and Llama-2-7B~\cite{touvron2023llama2}. We pretrain models for 300k steps, then evaluate performances on the MSRVTT, VATEX, AudioCaps, ClothoV2, COCO, and Flicker datasets for caption (CIDEr) and retrieval tasks (R@1).}
\vspace{-2mm}
\label{tab:arch}
\resizebox{0.98\linewidth}{!}{
\begin{tabular}{l cc cc cc}
\toprule
\multirow{3}{*}{Architecture} 
& \multicolumn{2}{c}{\color{VideoDark}{Video} }  
& \multicolumn{2}{c}{\color{AudioDark}{Audio} }
& \multicolumn{2}{c}{\color{ImageDark}{Image} }
\\\cmidrule{2-3} \cmidrule{4-5} \cmidrule{6-7}

 & {\color{VideoDark}MSRVTT}
 & {\color{VideoDark}VATEX}  
 & {\color{AudioDark}AudioCaps} 
  & {\color{AudioDark}ClothoV2} 
 & {\color{ImageDark}COCO}
 & {\color{ImageDark}Flickr} \\
 
 & CIDEr (\%) & R@1 (\%) & R@1 (\%) & CIDEr (\%) & R@1 (\%) & R@1 (\%)  \\ \midrule
 (i) Modality-Specific &74.3 &73.5 &42.3 & 22.3 & 65.2 & 88.4\\
 (ii) Text Encoder (BERT) & 77.0 & 53.2 & 23.1 & 43.9 & 46.7 & 51.6\\
 (iii) LLM (LLama-2-7B) & 75.2 & 60.3 & 14.7 & 43.6 & 60.8 & 81.3 \\
 (iv) ViT + LLM & \textbf{77.9} & \textbf{79.5} & \textbf{49.7} & \textbf{47.2} & \textbf{67.5} & \textbf{90.5}\\
 \bottomrule
\end{tabular}
}
\end{table*}

\textbf{Empirical Discovery.} Referring to Table~\ref{tab:arch}, we conclude several guidelines for designing architectures in omni-modal learning: \textbf{1}) \textit{Pure language models are difficult to retrieval tasks}. Both (ii) and (iii) deliver a significant performance drop in retrieval tasks. \textbf{2}) \textit{No more than 2 Encoders}. Comparing (i) with (ii) \& (iii), we observe that additional encoders are beneficial for retrieval tasks; however, comparison between (i) and (iv) suggests that discrepancies among multiple encoders can also hinder multimodal alignment. \textbf{3}) \textit{Language is an individual branch for alignment}. Comparing (ii) \& (iii), with (iv), improvements are significant in both retrieval and captioning.     

\subsection{Multimodal Context Construction}~\label{sec:method:modal}
\begin{figure}[t]
    \centering
    \includegraphics[width=1.0\linewidth]{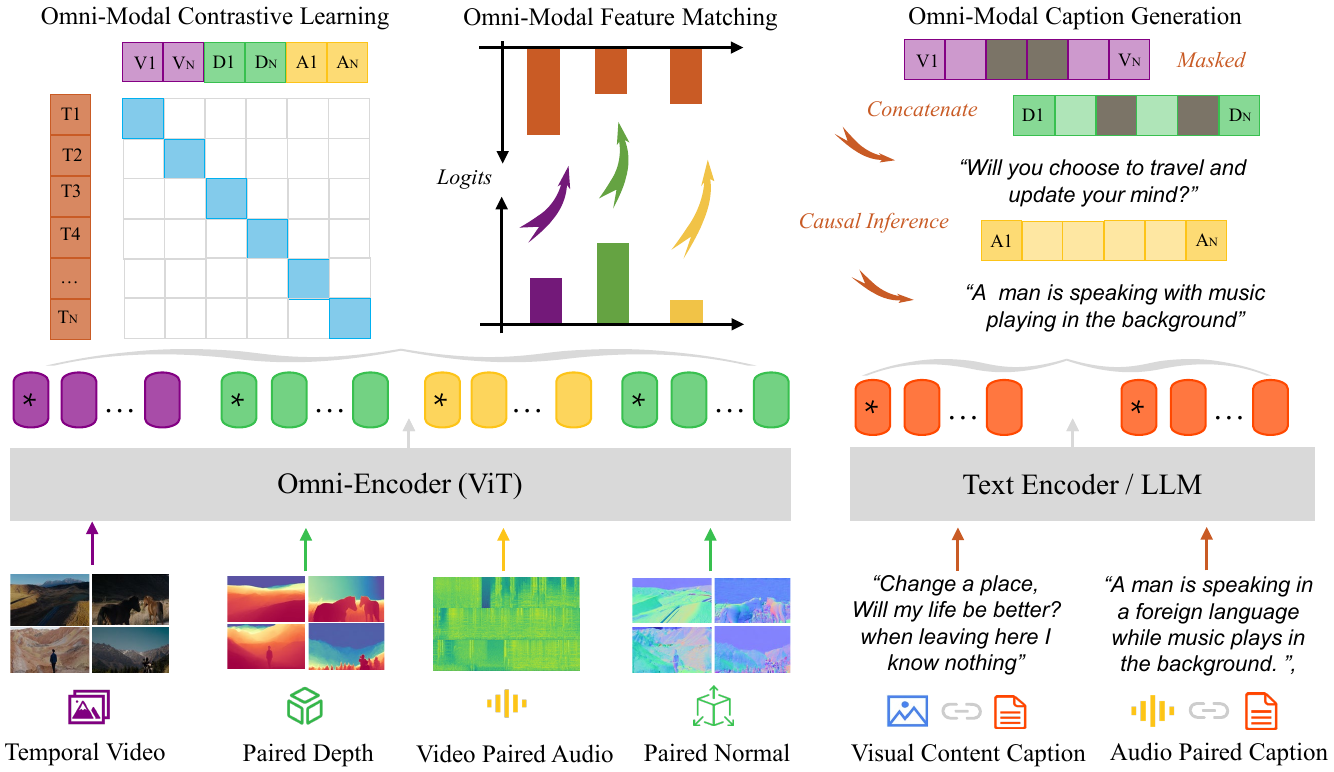}
    \caption{\textbf{Overview of Multimodal Context Pretraining Paradigm}. We use a shared ViT for multimodal feature extraction, and another branch is to employ a text encoder. We concatenate these multimodal sequences as multimodal contexts and perform contrastive learning and masked modeling.}
    \vspace{-4mm}
    \label{fig:framework}
\end{figure}

\textbf{Preliminary.} The context is proposed to assign a unique vector to each token in a sequence~\cite{vaswani2017attention}, which reinforces potential relevance between positions. Different modalities (\textit{e.g.}, text, image, audio) provide complementary information. Learning multimodal context leads to a more holistic and nuanced understanding of data. It can also leverage the strengths of each modality and guide the model to understand the interactions between different types of information. Therefore, we seek to construct the context relationship across diverse modalities and extend the learning capacity to omni-modalies. We provide the overview of \\mico pretraining paradigm in Figure~\ref{fig:framework}.

\textbf{Single Dataset with Multimodal Paired Data.} As mentioned in \S~\ref{sec:method:data}, we build a dataset with multimodal paired data \(\{(\boldsymbol{x}_I,\boldsymbol{x}_D, \boldsymbol{x}_N, \boldsymbol{x}_T^I), (\boldsymbol{x}_A,\boldsymbol{x}_T^A), (\boldsymbol{x}_V,\boldsymbol{x}_T^V)\}\), then we employ the omni-modal encoder \(f(\cdot;\theta)\) to extract features \(\boldsymbol{z}_I, \boldsymbol{z}_A, \boldsymbol{z}_V, \boldsymbol{z}_D, \text{ and } \boldsymbol{z}_N   \), then use text encoder to extract text features \(\boldsymbol{z}_T\). Therefore, we construct the context by a top-down design: \textbf{1}) For the whole multimodal embeddings, they share the same position embeddings \(\boldsymbol{E}_{\text{Pos}}\) to build a modality-fused context relationship across diverse modalities. \textbf{2}) Then, for each specific context, they're labeled by modality embeddings including \(\boldsymbol{E}^I_{\text{M}}, \boldsymbol{E}^A_{\text{M}}, \boldsymbol{E}^V_{\text{M}}, \boldsymbol{E}^D_{\text{M}}, \boldsymbol{E}^N_{\text{M}},\) \textit{etc} to indicate modality types. \textbf{3}) Within the same modality context, we employ the context embeddings \(\boldsymbol{E}^I_{\text{C}}\) to construct uni-modal context relationships. Thus, the construction of the multimodal context can be formulated as:
\begin{equation}
    \label{eq:single:mmcontext}
    \begin{aligned}
        \boldsymbol{z}_I &= [\boldsymbol{z}_I^1, \boldsymbol{z}_I^2, \cdots, \boldsymbol{z}_I^{L_I}] + \boldsymbol{E}_{\text{C}}^I, \hspace{2.5mm}\text{for each modality,} \\
    \boldsymbol{z} &= [\boldsymbol{z}_I + \boldsymbol{E}_{\text{M}}^I, \boldsymbol{z}_A + \boldsymbol{E}_{\text{M}}^A, \boldsymbol{z}_V + \boldsymbol{E}_{\text{M}}^V, \boldsymbol{z}_D + \boldsymbol{E}_{\text{M}}^D, \boldsymbol{z}_N + \boldsymbol{E}_{\text{M}}^N] + \boldsymbol{E}_{\text{Pos}},
    \end{aligned}
\end{equation}
where \(\boldsymbol{E}_{\text{C}}^I\) is up to the sample length of a specific modality. Meanwhile, the text features of specific captions can be easily concatenated, where their position embeddings \(\boldsymbol{E}_{\text{Pos}}^\prime\) are also shared:
\begin{equation}
    \label{eq:single:textcontext}
    \boldsymbol{z}_T = [\boldsymbol{z}_T^I, \boldsymbol{z}_T^A, \boldsymbol{z}_T^V] + \boldsymbol{E}_{\text{Pos}}^\prime.
\end{equation}

Please note that, in the context construction, we use vanilla position embedding~\cite{vaswani2017attention} to build these contexts in all of these learnable embeddings \(\boldsymbol{E}_{\text{Pos}}, \boldsymbol{E}_{\text{Pos}}^\prime, \boldsymbol{E}_{\text{M}}, \text{ and } \boldsymbol{E}_{\text{C}}\) instead of Rotary Position Embedding (RoPE)~\cite{su2024roformer}, which is similar to EVA-CLIP-8B~\cite{fang2023eva} and EVA-CLIP-18B.

\textbf{Multiple Datasets Combination of Cross-Modal Datasets.} Besides multimodal paired data, our proposed paradigm can also leverage existing web-scale text-image, text-audio, and text-video datasets to jointly pretraining models towards omni-modal universal representations. Given datasets \(\mathcal{D}_I = \{(\boldsymbol{x}_I^j, \boldsymbol{x}_T^j)\}_{j=1}^{N_I}, \mathcal{D}_A = \{(\boldsymbol{x}_A^j, \boldsymbol{x}_T^j)\}_{j=1}^{N_A}, \text{ and 
 }\mathcal{D}_V = \{(\boldsymbol{x}_V^j,\boldsymbol{x}_T^j)\}_{j=1}^{N_V}\), each pair of data possess local and simple context, for example, a pair of text-image data \((\boldsymbol{x}_I, \boldsymbol{x}_T)\) corresponds to a simple context \((\boldsymbol{z}_I+\boldsymbol{E}_{\text{Pos}}, \boldsymbol{E}_{\text{Pos}}^\prime)\), which may limit the learned representations of models. We propose to build the multimodal context by cross-dataset joint sampling with sampling context embedding \(\boldsymbol{E}_{\text{Sam}}\):
 \begin{equation}
     \label{eq:multi:mmcontext}
     \begin{aligned}
         (\boldsymbol{x}_I,\boldsymbol{x}^I_T) & = \mathtt{Sample}(\mathcal{D}_I), \hspace{1.0mm}(\boldsymbol{x}_A, \boldsymbol{x}^A_T) = \mathtt{Sample}(\mathcal{D}_A), \hspace{1.0mm}(\boldsymbol{x}_V, \boldsymbol{x}^V_T) = \mathtt{Sample}(\mathcal{D}_V), \\
         \boldsymbol{z}_I & = f(\boldsymbol{x}_I;\theta) + \boldsymbol{E}_{\text{Sam}}^{T-I}, \hspace{2mm}
         \boldsymbol{z}^I_T = f^\prime(\boldsymbol{x}_{T};\theta^\prime) +\boldsymbol{E}_{\text{Sam}}^{T-I}, \hspace{2mm} \text{for each modality,}\\
         \boldsymbol{z} &= [\boldsymbol{z}_I + \boldsymbol{E}_{\text{M}}^I, \boldsymbol{z}_A + \boldsymbol{E}_{\text{M}}^A, \boldsymbol{z}_V + \boldsymbol{E}_{\text{M}}^V] + \boldsymbol{E}_{\text{Pos}}, \hspace{1.5mm}
         \boldsymbol{z}_T = [\boldsymbol{z}_T^I, \boldsymbol{z}_T^A, \boldsymbol{z}_T^V] + \boldsymbol{E}_{\text{Pos}}^\prime.
     \end{aligned}
 \end{equation}
 In this way, we successfully combine existing multiple cross-modal datasets towards learning omni-modal universal representations by building more universal and complicated multimodal contexts (Equation~\ref{eq:multi:mmcontext}) for pretraining models, therefore, \textit{\\mico can outperform existing pretraining methods by better generalization learning ability, modality extensibility, and easier for scaling data}.

\subsection{Pretraining Objectives}~\label{sec:method:obj}
\textbf{\textit{Omni-modal Contrastive Learning.}} The omni-modality representations are denoted as \(\boldsymbol{z}\). Subsequently, \(\boldsymbol{z}\) and \(\boldsymbol{z}_T\) are projected into the same space using MLPs. The omni-modal contrastive learning is formulated by the dot product of \(\boldsymbol{z}\) and \(\boldsymbol{z}_T\). We use \(v^z\) and \(v^T\) to denote projected vectors:
\begin{equation}
\mathcal{L}_{\mathrm{Con}}=-\frac{1}{2} \sum_{i=1}^{N_B} \log \frac{\exp( \tau \cdot <v^z_{ i},  v^T_{i}>)}{\sum_{j=1}^{N_B}  \exp( \tau \cdot <v^z_{ i},  v^T_{j}>))}-\frac{1}{2} \sum_{i=1}^{N_B} \log \frac{\exp( \tau \cdot <v^z_{ i},  v^T_{i}>))}{\sum_{j=1}^{N_B} \exp( \tau \cdot <v^z_{ j},  v^T_{i}>))},
\end{equation}
where \(<\cdot,\cdot>\), $N_B$ and $\tau$ denote the dot product, batch size, and a learnable parameter.

\textbf{\textit{Omni-modal Feature Matching Process}} is designed to improve the semantic alignment between multimodal (knowledge modalities) and textual features. We employ an MLP layer to perform binary predictions \(p_v\) of \((\boldsymbol{z},\boldsymbol{z}_T)\). Following a hard negative mining strategy \cite{li2021align}, we assigns \(y=1\) if features are matched, and \(y=0\) otherwise.

\begin{equation}
\mathcal{L}_{\mathrm{Match}}=\mathbb{E}_{\left(v^z_{i}, v^T_{i}\right) \sim(\mathcal{Z},  \mathcal{T})}\left[y \log p_{v}+\left(1-y\right) \log \left(1-p_{v}\right)\right]
\end{equation}

\textbf{\textit{Omni-modal Caption Generation Process.}} We employ conditional causal masked (60\%) language modeling for generative omni-modal reasoning. In specific, a single-directional causal attention mask is used to avoid information leakage, and the masked tokens are reconstructed using a prediction layer of BERT~\cite{devlin2018bert}. We use $c_m$ and $c_{<m}$ to denote masked tokens and former tokens, respectively.
\begin{equation}
\mathcal{L}_{\mathrm{Gen}}=-\mathbb{E}_{\left(v^T_{i}, v^T_{i}\right) \sim(\mathcal{V},  \mathcal{T})} \log P\left(c_{m} \mid c_{<m}, v^z\right)
\end{equation}

\section{Experiment}~\label{sec:exp}
\vspace{-0.7cm}
\subsection{Experimental Setup}~\label{sec:exp:set}
We evaluate our model on three different benchmarks: \textbf{1}) Single-modality Understanding~\S~\ref{sec:exp:single} (following previous practices~\cite{radford2021learning,zhang2023meta,girdhar2023imagebind} in fine-tuning \& zero-shot setting in classification and forecasting tasks), \textbf{2}) Cross-modality Understanding~\S~\ref{sec:exp:cross} (following BEiT-3~\cite{wang2022image}, VAST~\cite{chen2023vast} in fine-tuning and dataset splits for Caption, QA, and retrieval tasks), and \textbf{3}) Multimodal Understanding with Large Language Models~\S~\ref{sec:exp:mllm} (following LLava~\cite{liu2023improvedllava}, VideoChat~\cite{li2023mvbench}, OneLLM~\cite{han2023onellm} in multimodal zero-shot QA). Detailed experimental settings including datasets introduction, splits, and evaluation metrics can be found in our Appendix~\ref{sec:app:data}.

\textbf{Implementation Details.} We implement our pretraining paradigm with ViT backbone scaling from ViT-B (8 NVIDIA Tesla A100 GPUs) to ViT-g (64 NVIDIA Tesla A100 GPUs). It takes about 2$\sim$6 days for pretraining. We pretrain models 200k steps for ViT-B, 300k steps for ViT-L and ViT-g. The initial learning rate is set to 1e-4, and a linear decay schedule is used. The batch
size on each GPU is set to 1,024. More implementation details can be found in the Appendix~\ref{sec:app:train}.
\begin{table*}[ht]
    \setlength{\tabcolsep}{4pt}
    \centering
    \vspace{-3mm}
    \caption{\textbf{State-of-the-art Abilities of \mico for Omni-modal Perception}. We conduct experiments on the single modality evaluation following the same practice of previous Sotas. We report the Accuracy (\%) of MMLU~\cite{hendrycks2020measuring}, IN-1K~\cite{deng2009imagenet}, K700~\cite{kay2017kinetics}, NYU-D~\cite{nyuv2}, Ego4D~\cite{grauman2022ego4d}, Indian Pines, and Fraud datasets, R@1 (\%) for MSR-VTT~\cite{xu2016msrvtt} and SYSU~\cite{wu2017rgb}, mAP for AS-2M~\cite{gemmeke2017audio}, F1-score for Fraud, and Mean Absulte Error\(\downarrow\) for PCQM4M and Global Weather Forecasting~\cite{wu2023interpretable} benchmarks.} 
    \vspace{-2mm}
    \resizebox{0.98\linewidth}{!}{
    \begin{tabular}{l | c c  c cc  c  c  c  c  c  c}
    \multirow{3}{*}{Methods (Backbone)}& \multicolumn{1}{c}{\color{ImageDark} \textChar{}} 
    & \multicolumn{1}{c}{\color{ImageDark} \imageChar{}} 
    & \multicolumn{1}{c}{\color{VideoDark} \videoChar{}} 
    & \multicolumn{1}{c}{\color{DepthDark} \depthChar{}} 
    & \multicolumn{1}{c}{\color{AudioDark} \audioChar{}}
    
    & \multicolumn{1}{c}{ \color{ThermalDark} \thermalChar{}} 
    & \multicolumn{1}{c}{\color{IMUDark} \imuChar{}}
    & \multicolumn{1}{c}{\color{GraphDark} \graphChar{}}
    & \multicolumn{1}{c}{\color{TimeDark} \timeChar{}}
    & \multicolumn{1}{c}{\color{HyperDark} \hyperChar{}}
     & \multicolumn{1}{c}{\color{TabularDark} \tabularChar{}}
    \\
    & {\color{TextDark} Text}
    & {\color{ImageDark} Image} 
    & {\color{VideoDark} Video} 
    & {\color{DepthDark} Depth} 
    & {\color{AudioDark} Audio} 
    & {\color{ThermalDark} Thermal}
    & {\color{IMUDark} IMU}
    & {\color{GraphDark} Graph}
    & {\color{TimeDark} Time-Series}
    & {\color{HyperDark} Hyperspectral}
    & {\color{TabularDark} Tabular}
    \\
    & {\color{TextDark} MMLU}
    & {\color{ImageDark} IN-1K} 
    & {\color{VideoDark} K700/MSR-VTT} 
    & {\color{DepthDark} NYU-D} 
    & {\color{AudioDark} AS-2M} 
    & {\color{ThermalDark} SYSU}
    & {\color{IMUDark} Ego4D}
    & {\color{GraphDark} PCQM4M}
    & {\color{TimeDark} Global Weather}
    & {\color{HyperDark} IP}
    & {\color{TabularDark} Fraud}
    \\
    \shline

    ImageBind (ViT-H)~\cite{girdhar2023imagebind} &
     43.6 %
    & 80.2
     & 42.9/36.8  %
    & 54.0 & 
    43.4 & 72.6 &25.0 & 0.815\(\downarrow\) %
    & 8.439\(\downarrow\) %
    & 83.6 %
    & 0.847 \\ %

     Meta-Trans (ViT-L)~\cite{zhang2023meta} & 37.3 & 88.1 & 33.2/31.5 & 41.5 & 38.9 & 71.3 & 73.9 & 0.886\(\downarrow\) & 7.892\(\downarrow\) & 78.1 & 0.809
    \\

    {\color{DarkGray} Absolute SOTA}
    & {\color{DarkGray} 90.0}~\cite{team2023gemini}
    & {\color{DarkGray} 91.0}~\cite{yu2022coca} %
    & {\color{DarkGray} 92.1/62.8}~\cite{wang2024internvideo2} %
    & {\color{DarkGray} 76.7}~\cite{girdhar2022omnivore} %
    & {\color{DarkGray} 48.6}~\cite{chen2022beats} %
    & {\color{DarkGray} 77.9~\cite{zhang2022modality} %
    }
    & {\color{DarkGray} 52.5~\cite{kazakos2021slow}}  %
    & {\color{DarkGray} 0.123~\cite{ying2021transformers}}

    & {\color{DarkGray} 7.602\(\downarrow\)~\cite{ding2023unireplknet}} %
    & {\color{DarkGray} 98.0~\cite{sigger2023diffspectralnet}} %
    & {\color{DarkGray} 0.860~\cite{padhi2021tabular}} %
    \\ \hline
    \mico (ViT-g)~\pub{Ours} &
    68.9 & 89.8 & 91.6/\textbf{64.3} & \textbf{84.6} & \textbf{50.5} & \textbf{80.3} & \textbf{77.2} & 0.742\(\downarrow\) & 7.834\(\downarrow\) & \textbf{98.5} & \textbf{0.913} \\
    \end{tabular}}
    \label{tab:emergent_zero_shot}
    \vspace{-7mm}
\end{table*}

\subsection{Evaluation on Single-modality Understanding }~\label{sec:exp:single}
\textbf{Exceptional Omni-modal Perception Abilities.} As shown in Table~\ref{tab:emergent_zero_shot}, \mico achieves state-of-the-art performances on a range of benchmarks across 10 modalities. For text understanding (MMLU), \mico attains the accuracy of 68.9\%, outperforming both ImageBind~\cite{girdhar2023imagebind} (43.6\%) and Meta-Transformer~\cite{zhang2023meta} (37.3\%). In image recognition (IN-1K), \mico delivers Top-1 Acc. of 89.8\%. On K700 and MSR-VTT, \mico achieves 91.6\% for Acc. and R@1 of 64.3\%, outperforming existing retrieval methods. Regrading 3D singe-view tasks (NYU-D), \mico outperforms the absolute SOTA~\cite{girdhar2022omnivore} by +7.9\%. On AS-2M, \mico achieves the mAP of 50.5\%, which is better than BEATS-3~\cite{chen2022beats} by +1.9\%. \mico also excels in thermal sensing (SYSU) and IMU tasks (Ego4D), \mico achieves an accuracy of 80.3\% and 77.2\%, respectively. 
\textit{These results highlight \mico's comprehensive and outstanding performances, establishing it as a powerful model for omni-modal perception.}
\begin{table}[ht]
\vspace{-4.5mm}
\caption{\textbf{Powerful Cross-Modal Abilities}. We evaluate \mico on the mainstream cross-modal tasks including 11 retrieval tasks (COCO~\cite{lin2014microsoft}, Flickr~\cite{plummer2015flickr30k}, ClothoV1~\cite{drossos2020clotho}, ClothoV2~\cite{drossos2020clotho}, AudioCaps~\cite{kim2019audiocaps}, MSRVTT~\cite{xu2016msr}, YouCook2~\cite{zhou2018towards}, VALOR-32K~\cite{chen2023valor}, VATEX~\cite{wang2019vatex}, DEDeMo~\cite{anne2017localizing}, and ANET~\cite{yu2019activitynet}), 7 caption tasks (COCO, ClothoV1, ClothoV2, AudioCaps, MSRVTT, YouCook2, VALOR-32K), and 6 question-answer (QA) tasks (TGIF~\cite{li2016tgif}, MSVD~\cite{xu2017video}, VQAv2~\cite{goyal2017making}, MSRVTT, MUSIC~ \cite{li2022learning}, and ANET) with the metrics of R@1, CIDEr, and Acc. Impressively,\textbf{ \mico archives 20 new SoTA performances}. }
\label{tab:cross_modal}
\centering
\vspace{1.5mm}
\resizebox{0.98\linewidth}{!}{
\begin{tabular}{l}
\begin{tabular}
{l@{\hskip 3.3mm}c@{\hskip 3.3mm}c@{\hskip 3.3mm}c@{\hskip 3.3mm}c@{\hskip 3.3mm}c@{\hskip 3.3mm}c@{\hskip 3.3mm}c}
\multirow{2}{*}{\begin{tabular}[c]{@{}c@{}} \color{ImageDark} \imageChar{}\\ {\color{ImageDark}Image}\end{tabular}}

& \multicolumn{3}{c}{{\color{ImageDark} Text-to-Image Retrieval}} & {\color{ImageDark} Image Caption} & \multicolumn{3}{c}{{\color{ImageDark} Visual QA}} \\
\cmidrule(lr){2-4} \cmidrule(lr){5-5} \cmidrule(lr){6-8}
& {COCO}         & {Flickr}         & {Flickr(ZS)}          & {COCO}         & {TGIF} & {MSVD}  & {VQAv2}  \\
\midrule

{\color{DarkGray} SOTA}                       &    {\color{DarkGray}{68.3}}~\cite{li2023blip}           &        {\color{DarkGray}90.3}~\cite{wang2022image}         &      {\color{DarkGray}89.7}~\cite{li2023blip}         &       {\color{DarkGray}{154.9}}*~\cite{wang2022ofa}        &    {\color{DarkGray}78.7}~\cite{chen2023valor}   &   {\color{DarkGray}60.2}~\cite{kuo2023mammut}    & {\color{DarkGray}{84.3}}~\cite{chen2022pali}       \\
\mico
& 68.1
& \reshl{91.1}{0.8}
& \reshl{90.1}{0.4}
& 152.4
& \reshl{78.9}{0.2}
& \reshl{60.4}{0.2} 
& 80.5
\\
\end{tabular}
\\\\
\begin{tabular}
{l@{\hskip 5mm}c@{\hskip 5.2mm}c@{\hskip 5.2mm}c@{\hskip 5.2mm}@{\hskip 5.2mm}c@{\hskip 5.2mm}c@{\hskip 5.2mm}c}
\multirow{2}{*}{\begin{tabular}[c]{@{}c@{}}\color{AudioDark} \audioChar{}\\ {\color{AudioDark} Audio}\end{tabular}} & \multicolumn{3}{c}{{\color{AudioDark} Text-to-Audio Retrieval}} & \multicolumn{3}{c}{{\color{AudioDark} Audio Caption}} \\
\cmidrule(lr){2-4} \cmidrule(lr){5-7} 
& {ClothoV1}            & {ClothoV2}      & {AudioCaps}     &  {ClothoV1} & {ClothoV2}      & {AudioCaps}     \\
\midrule

{\color{DarkGray} SOTA}                       & {\color{DarkGray}17.5}~\cite{chen2023valor}  &   {\color{DarkGray}21.5}~\cite{mei2023wavcaps}        & {\color{DarkGray}42.2}~\cite{mei2023wavcaps}          &    {\color{DarkGray}42.3}~\cite{chen2023valor} &  {\color{DarkGray}48.8}~\cite{mei2023wavcaps}             &  {\color{DarkGray}{78.7}}~\cite{mei2023wavcaps}  \\

\mico 
& \reshl{21.2}{3.7}
& \reshl{23.3}{1.8}
& 41.0
& \reshl{49.6}{7.3}
& \reshl{50.8}{2.0}
& 66.2 \\
\end{tabular}
\\ \\

\begin{tabular}
{l@{\hskip 3mm}c@{\hskip 4mm}c@{\hskip 4mm}c@{\hskip 3mm}@{\hskip 3mm}c@{\hskip 4mm}c@{\hskip 3mm}c}
\multirow{2}{*}{\begin{tabular}[c]{@{}c@{}} \color{VideoDark} \videoChar{} \color{AudioDark} \audioChar{}\\ {{\color{VideoDark} Video}-{\color{AudioDark} Audio}}\end{tabular}} & \multicolumn{6}{c}{{\color{VideoDark}Text-to-Video}-{\color{AudioDark}Audio}  {\color{VideoDark}Retrieval}}                                 \\
\cmidrule(lr){2-7}
& {MSRVTT} & {YouCook2} & {VALOR-32K} & {VATEX} & {DiDeMo} & {ANET} \\
\midrule

{\color{DarkGray} SOTA}                         & 
{\color{DarkGray}54.4}~\cite{chen2023valor}       &   {\color{DarkGray}31.3}~\cite{li2021value}       &       {\color{DarkGray}73.2}~\cite{chen2023valor}    &      {\color{DarkGray}76.9}~\cite{chen2023valor} &    {\color{DarkGray}57.6}~\cite{chen2023valor}    &    {\color{DarkGray}63.4}~\cite{chen2023valor} \\

\mico
& \reshl{64.3}{9.9}
& \reshl{51.3}{20.0}
& \reshl{78.7}{5.5}
& \reshl{81.3}{4.4} 
& \reshl{63.6}{6.0}
& \reshl{68.5}{5.1}    \\
\end{tabular}
\\\\
\begin{tabular}
{l@{\hskip 3mm}c@{\hskip 4mm}c@{\hskip 4mm}c@{\hskip 3mm}@{\hskip 3mm}c@{\hskip 4mm}c@{\hskip 3mm}c}
\multirow{2}{*}{\begin{tabular}[c]{@{}c@{}} \color{VideoDark} \videoChar{} \color{AudioDark} \audioChar{} \\

{{\color{VideoDark} Video}-{\color{AudioDark} Audio}}\end{tabular}} 
& \multicolumn{3}{c}{{\color{VideoDark} Video}-{\color{AudioDark} Audio} {\color{VideoDark} Caption}}              
& \multicolumn{3}{c}{{\color{VideoDark} Video}-{\color{AudioDark} Audio} {\color{VideoDark} QA}} \\
\cmidrule(lr){2-4} \cmidrule(lr){5-7}
& {MSRVTT} & {YouCook2} & {VALOR-32K} 
& {MSRVTT}  & {MUSIC} & {ANET} \\
\midrule
{\color{DarkGray} SOTA}                         &   
{\color{DarkGray}74.0}~\cite{chen2023valor}     &  {\color{DarkGray}190.0}~\cite{ko2023meltr}        & {\color{DarkGray}61.5}~\cite{chen2023valor}      
&  {\color{DarkGray}49.2}~\cite{chen2023valor}       &   {\color{DarkGray}78.9}~\cite{chen2023valor}    &{\color{DarkGray}48.6}~\cite{chen2023valor}     \\

\mico 
& \reshl{79.3}{5.3} 
& \reshl{197.8}{7.8}
& \reshl{62.8}{1.3}  
& \reshl{50.4}{1.2} 
& \reshl{79.7}{0.8} 
& \reshl{51.0}{2.4}  \\
\end{tabular}
\vspace{-4mm}
\end{tabular}
}
\vspace{-2mm}
\end{table}
\subsection{Evaluation on Cross-Modal Understanding }~\label{sec:exp:cross}
Table~\ref{tab:cross_modal} illustrates the powerful performances of \mico on 25 cross-modal benchmarks, \textbf{achieving more than 20 new SOTA performances}. For text-to-image retrieval, \mico achieves outstanding results with R@1 of 68.1\% on COCO, and 91.1\% on Flickr, outperforming previous SOTA methods. For VQA, \mico demonstrates robust performance with accuracy scores of 78.9\% on TGIF, 60.4\% on MSVD, and 80.5\% on VQA v2, highlighting its strong visual comprehension and reasoning abilities. In text-to-audio retrieval, \mico achieves outstanding performances of 21.2\% on ClothoV1, 23.3\% on ClothoV2, and 41.0\% on AudioCaps, while in audio captioning, it achieves 49.6\% on ClothoV1, and 50.8\% on ClothoV2, all outperforming previous best results. For text-to-video retrieval, \mico sets new SOTA performances with metrics of 64.3\% R@1 on MSRVTT and 81.3\% on VATEX, and in video-audio caption, it achieves impressive performances of 79.3\% on MSRVTT, 197.8\% on YouCook2, and 62.8\% on VALOR-32K. Finally, in video-audio QA, \mico also delivers superior performances of 50.4\% on MSRVTT, 79.9\% on MUSIC, and 51.0 on ANET. \textit{These results collectively highlight \mico's exceptional and versatile capabilities in cross-modal comprehension and reasoning tasks, establishing it as a promising direction in this field.}

\begin{table*}[t]
\centering
\caption{\textbf{Evaluation on LLM Benchmarks.} The MLLM evaluation involves 6 VQA tasks (GQA~\cite{hudson2019gqa}, VQAv2~\cite{goyal2017vqav2}, OKVQA~\cite{okvqa}, TextVQA (TVQA)~\cite{singh2019textvqa}, ScienceQA (SQA)~\cite{lu2022learn} and Vizwiz~\cite{gurari2018vizwiz}), 2 image captioning tasks (Nocaps~\cite{agrawal2019nocaps} and Flickr30K~\cite{plummer2015flickr30k}), and 4 multimodal benchmarks (MME~\cite{fu2023mme}, MM Bench (MMB)~\cite{liu2023mmbench}, MMVet~\cite{yu2023mmvet} and SEED~\cite{li2023seed}). The LLMs are Chinchilla~\cite{hoffmann2022training}, Vicuna~\cite{vicuna}, Qwen~\cite{bai2023qwen_llm}, LLaMA~\cite{touvron2023llama} and LLaMA2~\cite{touvron2023llama2}. The evaluation metrics for VQA and captioning tasks are accuracy and CIDEr, respectively.
The results in \textbf{bold} and \underline{underline} are the best and second-best results, respectively.
}
\vspace{-2mm}
\resizebox{\textwidth}{!}{%
\begin{tabular}{lccccccccccccc}
\toprule
\multirow{2}{*}{Method}                        & \multirow{2}{*}{LLM} & \multicolumn{6}{c}{\color{ImageDark} Visual Question Answering}                     & \multicolumn{2}{c}{ \color{ImageDark} Image Caption} & \multicolumn{4}{c}{ \color{ImageDark} MM Benchmark} \\
\cline{3-8} \cline{9-10} \cline{11-14}
&                      & GQA   & VQAv2 & OKVQA & TVQA & SQA  & Vizwiz & NoCaps          & Flickr           & MME    & MMB  & MMVet & SEED \\
\midrule
\multicolumn{4}{l}{\textbf{\textit{Vision Specialist LLM}}}                                                \\
Flamingo-9B~\cite{alayrac2022flamingo}        & Chinchilla-7B        & -     & 51.8  & 44.7  & 30.1 & -    & 28.8   & -               & 61.5             & -      & -    & -     & -    \\
Flamingo-80B~\cite{alayrac2022flamingo}       & Chinchilla-70B       & -     & 56.3  & 50.6  & 31.8 & -    & 31.6   & -               & 67.2             & -      & -    & -     & -    \\
BLIP-2~\cite{li2023blip}                      & Vicuna-7B            & -     & -     & -     & 40.1 & 53.8 & -      & 107.5           & 74.9             & -      & -    & -     & -    \\
BLIP-2~\cite{li2023blip}                      & Vicuna-13B           & 41.0  & 41.0  & -     & 42.5 & 61   & 19.6   & 103.9           & 71.6             & 1293.8 & -    & 22.4  & -    \\
InstructBLIP~\cite{instructblip}              & Vicuna-7B            & 49.2  & -     & -     & 50.1 & 60.5 & 34.5   & 123.1           & 82.4             & -      & 36   & 26.2  & - \\
InstructBLIP~\cite{instructblip}              & Vicuna-13B           & 49.5  & -     & -     & 50.7 & 63.1 & 34.3   & 121.9           & 82.8             & 1212.8 & -    & 25.6  & - \\
IDEFICS-9B~\cite{laurenccon2023obelisc}       & LLaMA-7B             & 38.4  & 50.9  & 38.4  & 25.9 & -    & 35.5   & -               & 27.3             & -      & 48.2 & -     & -    \\
IDEFICS-80B~\cite{laurenccon2023obelisc}      & LLaMA-65B            & 45.2  & 60.0  & 45.2  & 30.9 & -    & 36.0   & -               & 53.7             & -      & 54.5 & -     & -    \\
LLaMA-Ad.v2~\cite{gao2023llama}               & LLaMA-7B             & 43.9  & -     & 55.9  & 43.8 & 54.2 & -      & 42.7            & 30.5             & 972.7  & 38.9 & \textbf{31.4}  & 32.7 \\
Qwen-VL~\cite{bai2023qwen}                    & Qwen-7B              & 57.5  & 78.2  & 56.6  & \textbf{61.5} & 68.2 & 38.9   & 120.2           & 81.0             & \underline{1487.5} & 60.6 & -     & 58.2 \\
LLaVA-v1.5~\cite{liu2023improvedllava}        & Vicuna-7B            & \textbf{62.0}  & \textbf{78.5}  & -     & \underline{58.2} & 66.8 & \textbf{50.0}   & -               & -                & \textbf{1510.7} & \underline{64.3} & \underline{30.5}  & 58.6 \\
\midrule                                         
\multicolumn{4}{l}{\textbf{\textit{Multimodal Generalist LLM}}}  \\
ImageBind-LLM~\cite{han2023imagebind}         & LLaMA-7B             & 41.1  & -     & -     & 24.0 & 51.4 & -      & 29.6            & 23.5             & 775.7  & -    & -     & -    \\
ChatBridge-13B~\cite{zhao2023chatbridge}      & Vicuna-13B           & {41.8}  & -     & {45.2}  & -    & -    & -      & \underline{115.7}           & \textbf{82.5}             & -      & -    & -     & -    \\
AnyMAL-13B~\cite{moon2023anymal}              & LLaMA2-13B           & -     & 59.6  & 33.1  & 24.7 & 52.7 & 24.4   & -               & -                & -      & -    & -     & -    \\
AnyMAL-70B~\cite{moon2023anymal}              & LLaMA2-70B           & -     & {64.2}  & 42.6  & {32.9} & \underline{70.8} & {33.8}   & -               & -                & -      & -    & -     & -    \\
{OneLLM-7B }~\pub{CVPR'24}                     & {LLaMA2-7B}   & 59.5  & 71.6 & \textbf{58.9}  & 34.0 & 63.4 & 45.9   & 115.9           & 78.6             & 1392.0 & 60.0 & 29.1  & \underline{61.2} \\
\midrule
\rowcolor{Gray}
\textbf{\mico-Chat-7B}                     & \textbf{Vicuna-7B}   & \underline{61.5}  & \underline{78.1}  & \underline{56.6}  & {53.4} & \textbf{71.3} & \underline{49.1}   & {111.8}           & {76.3}             & 1485.7 & \textbf{65.2} & \textbf{31.4}  & \textbf{67.7} \\
\bottomrule
\end{tabular}%
}
\vspace{-4mm}
\label{tab:image_eval}
\end{table*}

\begin{table}[t]
\centering
\captionof{table}{\textbf{Zero-Shot Audio \& Video generative benchmark with LLMs.} We evaluate models by audio captioning on Clotho Caption~\cite{drossos2020clotho}, audio QA on Clotho AQA~\cite{lipping2022clotho} and VideoChatGPT scoring benchmark~\cite{maaz2023video} using the same Vicuna-7B.}
\label{tab:mllm_eval}
\resizebox{0.46\linewidth}{!}{%
\begin{tabular}{lcccc}
\toprule
\multirow{2}{*}{Method}              & \multirow{2}{*}{0-shot}    & \multicolumn{2}{c}{\color{AudioDark} Clotho Caption} & { \color{AudioDark} Clotho AQA} \\
&                            & CIDEr           & SPIDEr           & Acc.       \\
\midrule

FeatureCut~\cite{ye2022featurecut}  & \ding{55}                  & 43.6            & 27.9             & -          \\
Wavcaps~\cite{mei2023wavcaps}       & \ding{55}                  & 48.8            & 31.0             & -          \\
MWAFM~\cite{li2023multi}            & \ding{55}                  & -               & -                & 22.2       \\
Pengi~\cite{deshmukh2023pengi}      & \ding{55}                  & -               & 27.1             & 64.5       \\
\midrule
ChatBridge-13B~\cite{zhao2023chatbridge}& \ding{51}              & 26.2            & -                & -          \\
OneLLM-7B           & \ding{51}                  & {29.1}   & {19.5}             & {57.9}       \\
\midrule
\rowcolor{Gray}
\textbf{\mico-Chat-7B}~\pub{Ours}           & \ding{51}                  & \textbf{33.3}   & \textbf{21.9}             & \textbf{63.9}       \\
\bottomrule
\end{tabular}%
}
\resizebox{0.48\linewidth}{!}{
\begin{tabular}{lccccc}
\toprule
\multirow{2}{*}{Method}  & {\color{VideoDark} Cor.}          & {\color{VideoDark} Det.}     & {\color{VideoDark} Con.} & {\color{VideoDark} Tem.} & {\color{VideoDark} Cons.}                             \\            \\
\midrule
VideoLLaMA~\cite{zhang2023videollama} & 1.96 & 2.18 & 2.16 & 1.82 & 1.79 \\
VideoChat~\cite{li2023videochat} & 2.23 & 2.50 & 2.53 & 1.94 & 2.24\\
Video-ChatGPT~\cite{maaz2023video} & 2.40 & 2.52 & 2.62 & 1.98 & 2.37 \\
BT-Adapter~\cite{liu2023one} & 2.68 & 2.69 & 3.27 & 2.34 & 2.46 \\
LLaMa-VID~\cite{li2023llama} & 2.96 & 3.00 & 3.53 & 2.46 & 2.51 \\
\midrule
\rowcolor{Gray}
\textbf{\mico-Chat-7B}~\pub{Ours}                 & \textbf{3.00} & \textbf{3.01} & \textbf{3.61} & \textbf{2.49} & \textbf{2.71}     \\
\bottomrule
\end{tabular}%
}
\vspace{-1mm}
\end{table}

\begin{table*}[t!]
 \centering
 \caption{\textbf{Zero-shot Video QA with LLMs}. In comparison with leading methods, we report results with 1 token for each frame, where Res. indicates image resolution. }
 \label{tab:mllm_video}
 \vspace{-2mm}
 \resizebox{0.98\linewidth}{!}{
 \begin{tabular}{llcccccccccccc}
  \toprule
  \multirow{2}{*}{Method} & \multirow{2}{*}{LLM} & \multirow{2}{*}{Res.} & \multicolumn{2}{c}{\bf {\color{VideoDark} MSVD-QA}} & & \multicolumn{2}{c}{\bf {\color{VideoDark} MSRVTT-QA}} & & \multicolumn{2}{c}{\bf {\color{VideoDark} ActivityNet-QA}} \\ \cline{4-5} \cline{7-8} \cline{10-11}
  & & & Acc & Score & & Acc & Score & & Acc & Score \\
  \midrule
  FrozenBiLM~\cite{yang2022zero} & DeBERTa-V2 & 224 & 32.2 & -- & & 16.8 & -- & & 24.7 & -- \\
  VideoLLaMA~\cite{zhang2023video} & Vicuna-7B & 224 & 51.6 & 2.5 & & 29.6 & 1.8 & & 12.4 & 1.1 \\
  LLaMA-Adapter~\cite{gao2023llama} & LLaMA-7B & 224 & 54.9 & 3.1 & & 43.8 & 2.7 & & 34.2 & 2.7 \\
  VideoChat~\cite{li2023videochat} & Vicuna-7B & 224 & 56.3 & 2.8 & & 45.0 & 2.5 & & 26.5 & 2.2 \\
  Video-ChatGPT~\cite{maaz2023video} & Vicuna-7B & 224 & 64.9 & \underline{3.3} & & 49.3 & 2.8 & & 35.2 & 2.7 \\ 
  LLaMA-VID~\cite{li2023llama} & Vicuna-7B & 224 & \underline{69.7} & { 3.7} & & \underline{57.7} & \underline{3.2} & & \underline{47.4} & {3.3} \\VideoChat2~\cite{li2023mvbench}~\pub{CVPR'24} & Vicuna-7B & 224 &  70.0 & 3.9 
  && 54.1 & 3.3 
  && 49.1 & 3.3 \\
  \midrule
  \rowcolor{gray!20}
  \mico-Chat-7B & Vicuna-7B & 224 &  \textbf{73.7} & \textbf{4.1} & & \textbf{60.1} & \textbf{3.6} & & \textbf{50.1} & \textbf{3.3}\\
  \bottomrule
\end{tabular}
 }
 \vspace{-2mm}
\end{table*}
\vspace{-1mm}
\subsection{Evaluation on Multimodal Understanding with Large Language Models}~\label{sec:exp:mllm}
\textbf{\mico highlights its Omni-modal Zero-shot Comprehension and Reasoning Abilities}. Beyond traditional caption, retrieval, and QA tasks, we also evaluate the abilities of \mico aligned with LLMs for zero-shot multimodal QA. We use ChatBridge~\cite{zhao2023chatbridge} as our baseline and Vicuna-7B as the large language model for each modality. As shown in Table~\ref{tab:image_eval}, \ref{tab:mllm_eval}, and \ref{tab:mllm_video}, \mico-Chat-7B shows outstanding performances across both Vision LLMs and Multimodal LLMs. It directly delivers outstanding performances on the SQA (71.3\%), MMB (65.2\%), MMVet (31.4\%), and SEED (67.7\%) benchmarks while another 4 competitive performances. Besides, \mico-Chat-7B also delivers significantly impressive performances on both zero-shot caption and QA tasks on audio and video modalities, where \textbf{\mico-Chat-7B achieves 6 new SOTA performances} including Clotho Caption, AQA, MSVD-QA, MSRVTT-QA, ActivityNet-QA. \textit{These results are important proof that the \mico pretraining paradigm shows a promising direction in developing large omni-modal models.}   

\begin{table*}[t]
\vspace{-2mm}
\caption{\textbf{Ablation Study} on pretraining modalities, data scale, pretraining process, and parameters. Our default setting is to pretrain a base model for 30k steps with 10M data using all objective functions and evaluate it on the MSRVTT, VATEX, DIDEMO, MSVD, AudioCaps, ClothoV2, COCO, and Flicker datasets for retrieval tasks.}
\label{tab:ablation}
\centering
\resizebox{0.98\linewidth}{!}{
\begin{tabular}{llccccccccc}
\toprule
\multirow{3}{*}{Model} &\multirow{3}{*}{Factors} &
\multicolumn{4}{c}{\color{VideoDark}{Video} \videoChar{}}  
& \multicolumn{2}{c}{\color{AudioDark}{Audio} \audioChar{}}
& \multicolumn{2}{c}{\color{ImageDark}{Image} \imageChar{}}
& Average \\

\cmidrule (lr){3-6} \cmidrule (lr){7-8} \cmidrule (lr){9-10}
 &  
 & \multicolumn{1}{c}{MSRVTT({\color{VideoDark}V}{\color{AudioDark}A})}    
 & \multicolumn{1}{c}{VATEX({\color{VideoDark}V}{\color{AudioDark}A})}
 & \multicolumn{1}{c}{DIDEMO({\color{VideoDark}V}{\color{AudioDark}A})} 
 & \multicolumn{1}{c}{MSVD({\color{VideoDark}V})}  
 & \multicolumn{1}{c}{AudioCaps({\color{AudioDark}A})} 
  & \multicolumn{1}{c}{ClothoV2({\color{AudioDark}A})} 
 & COCO({\color{ImageDark}I}) & Flickr({\color{ImageDark}I})
 \\
\midrule
\multicolumn{2}{l}{\textbf{Pretraining Modalities}} \\

(a) &I   
& 39.7 & 57.3 & 38.4 & 39.7 & 10.2 & 4.4 & 50.2 & 75.7 & 39.4\\
(b) &I+3D
& 42.0 & 58.5 & 38.1 & 40.1 & 10.8 & 4.2 & 51.2 & 76.9 & 40.2\\
(c) &I+A
& 37.6 & 56.2 & 30.8 & 36.2 & 22.0 & 14.5 & 46.8 & 71.0 & 39.4 \\
(d) &I+V
& 41.7 & 60.9 & 39.2 & 42.6 & 12.2 & 5.1 & 51.3 & 77.0 & 41.3\\
(e) &I+V+A
& 42.2 & 61.1 & 40.1 & 41.2 & 23.4 & 15.4 & 48.7 & 74.2 & 43.2\\
(f) &I+V+A+3D
& \reshl{45.7}{6.0} & \reshl{64.0}{6.7} & \reshl{42.7}{4.3} & \reshl{42.8}{3.1} & \reshl{24.6}{14.4} & \reshl{15.9}{11.5} & 49.9 & \reshl{77.1}{1.4} & \reshl{45.3}{5.9} \\
\midrule

\multicolumn{2}{l}{\textbf{Data Scale}} \\
(h) & 1M & 44.2 & 63.2 & 40.1 & 40.7 & 21.9 & 11.2 & 48.2 & 77.5  & 43.4\\
(i) & 10M & 45.7 & 64.0 & 42.7 & 42.8 & 24.6 & 15.9 & 49.9 & 77.1  & 45.3\\
(j) & 110M & 48.5 & 65.7 & 41.7 & 43.0 & 26.3 & 17.1 & 49.6 & 78.1  & 46.3\\
(k) & 334M & \reshl{49.1}{4.9} & \reshl{66.3}{3.1} & \reshl{43.2}{3.1} & \reshl{44.1}{3.4} & \reshl{27.0}{5.1} & \reshl{17.5}{6.3} & \reshl{51.5}{3.3} & \reshl{80.9}{3.4}  & \reshl{47.5}{4.1}\\
\midrule

\multicolumn{2}{l}{\textbf{Pretraining Process}} \\
(l) & \(\mathcal{L}_\text{Con}\) & 40.1 & 57.4 & 39.1 & 41.4 & 23.1 & 14.4 & 47.4 & 73.7 & 42.1 \\
(m) & \(\mathcal{L}_\text{Con} + \mathcal{L}_\text{Match}\) & 43.9 & 61.4 & 38.0 & 41.6 & 23.6 & 15.5 & 48.8 & 74.3 & 43.4 \\
(n) & \(\mathcal{L}_\text{Con} + \mathcal{L}_\text{Match} + \mathcal{L}_\text{Gen}\) & \reshl{45.7}{5.6} & \reshl{64.0}{6.6} & \reshl{42.7}{3.6} & \reshl{42.8}{1.4} & \reshl{24.6}{1.5} & \reshl{15.9}{1.5} & \reshl{49.9}{2.5} & \reshl{77.1}{3.4} & \reshl{45.3}{3.2}\\
\midrule
\multicolumn{2}{l}{\textbf{Model Scale}} \\
(o) & Base-86M & 45.7 & 64.0 & 42.7 & 42.8 & 24.6 & 15.9 & 49.9 & 77.1  & 45.3\\
(p) & Large-331M &  58.2 & 72.0 & 57.2 & 52.8 & 31.6 & 18.7 & 60.8 & 87.5 & 54.9\\
(q) & Giant-1.3B & \reshl{62.5}{16.8} & \reshl{79.9}{15.9} & \reshl{61.1}{18.4} & \reshl{56.0}{13.2} & \reshl{37.4}{12.8} & \reshl{20.8}{4.9} & \reshl{67.1}{17.2} & \reshl{90.7}{13.6} & \reshl{59.4}{14.1}\\
\midrule
\end{tabular}
}
\vspace{-4mm}
\end{table*}

\begin{figure}[ht]
    \centering
    \includegraphics[width=0.98\linewidth]{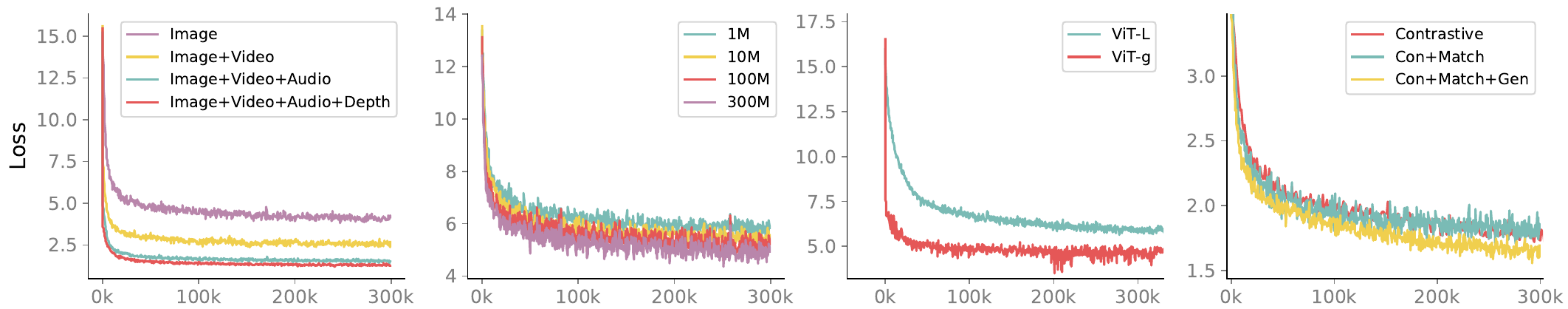}
    \caption{\textbf{Multimodal Scaling Law}. Training loss curves for \mico under scaling factors (modality, data, parameters, process) settings.}
    \vspace{-3mm}
    \label{fig:scaling_law}
\end{figure}

\subsection{Ablation Study: Multimodal Scaling Laws}~\label{sec:exp:ablate}
\textbf{Scaling Modalities}. From (a) to (f), we gradually scale up input modalities. In Figure~\ref{fig:scaling_law}, all modalities (I+V+A+3D) achieves the highest scores, highlighting the importance and effectiveness of \mico for diverse multimodal inputs.

\textbf{Scaling Multimodal Data.} From (h) to (k) in Table~\ref{tab:ablation}, we investigate the impact of the omni-modal data scale from 1M to 334M. It proves that the \mico has great potential for further scaling.

\textbf{Pretraining Objectives.} From (l) to (n), we analyze the impact of each pretraining objective. The combination of contrastive, matching, and generative losses (\(\mathcal{L}_\text{Con} + \mathcal{L}_\text{Match} + \mathcal{L}_\text{Gen}\)) yields the best performance, demonstrating the value of multiple complementary objectives.

\textbf{Scaling Parameters.} From (o) to (q), we assess the effect of model size. Larger models, particularly the Giant-1.3B, show superior performance, confirming that increasing model parameters with \mico enhances learning and generalization abilities across diverse modalities.

\section{Conclusion and Limitation}~\label{sec:conclusion}
In this paper, we propose a novel framework, termed \mico, to train foundation models with enhanced visual perception abilities and omni-modal capacities. With experiments on a reasonably large scale of both model and data, we conclude that the key to omni-modal learning is to simulate the multimedia cognition process of the human brain. In \mico, we use image, depth, and normal maps to simulate the fundamental visual perception ability, distance spatial awareness, and geometry awareness of human visual cognition. In addition, captions, audio, and video provide prior knowledge, auditory perception, and spatial-temporal awareness. In future work, we plan to enhance our joint pretraining by incorporating additional modalities, including optical flow, IMU data, and event files, \textit{etc}. We believe \mico is an important attempt to simulate the multimedia cognition of human brains, and we expect it could inspire future works to develop more powerful omni-modal foundation models.


\bibliographystyle{unsrt}
\bibliography{egbib}


\clearpage

\appendix

{\Large\textbf{Appendix}}

\section{Summary}
The appendix is organized as follows:
\begin{itemize}
	\item \S~\ref{sec:app:train} detailed training and evaluation settings of our models including hyper-parameters regarding models and optimizers.
	\item \S~\ref{sec:app:data} presents a comprehensive introduction on the datasets we use for evaluation and their corresponding metrics.
\end{itemize}

\section{Training Configuration}~\label{sec:app:train}

\subsection{Pretraining Settings}

We detail the specific pretraining configurations of \mico, focusing on the multi-dataset joint training corpora, the dataset mix ratios for each corpus, and the learning objectives for each corpus. To improve data quality, we employed a trained vision captioner to generate new captions for the CC4M datasets, replacing the original captions. Although \mico has only been trained for 300,000 steps, it has already demonstrated outstanding performance on various downstream tasks. We anticipate that further increasing the number of training steps will significantly enhance the model's capabilities.

The pretraining of \mico involves a combination of different datasets, each contributing uniquely to the model's learning process. The model, with a parameter size of 1.0 billion and a sample size of 334 million, utilizes a diverse training corpus to achieve its results.

1. VAST-27M: This dataset contributes 324 million samples to the training process. With a batch size of 2048, the model undergoes 160,000 steps, completing one epoch.

2. VALOR-1M: In this dataset, 1 million samples are used with a batch size of 1024. The training spans 70,000 steps, which equates to approximately 71.7 epochs.

3. WavCaps, CC4M, and WebVid-2.5M: These datasets are combined, contributing 9 million samples in total. The batch size for this combined dataset is 1024, and the model is trained over 70,000 steps, resulting in 8.0 epochs.

The careful selection and combination of these datasets, along with the application of new, high-quality captions for the CC4M datasets, enhance the training efficiency and the quality of the learned representations.

\subsection{Fine-tuning Settings}

We detail the downstream task finetuning settings, specifying the learning rate, batch size, epoch, training objectives, and resolution. The configurations also include the number of sampled video frames or audio clips used in training and testing phases. Here are the comprehensive settings:

\textbf{Retrieval Tasks (RET)}
\begin{itemize}
    \item \textbf{Image-Text Modality}
    \begin{itemize}
        \item \textbf{MSCOCO}: Learning rate of 1e-5, batch size of 256, 5 epochs, with the objective for retrieval, and a resolution of 384.
        \item \textbf{Flickr}: Learning rate of 1e-5, batch size of 256, 5 epochs, with the objective for retrieval, and a resolution of 384.
    \end{itemize}
    \item \textbf{Audio-Text Modality (A-T)}
    \begin{itemize}
        \item \textbf{ClothoV1/V2}: Learning rate of 2e-5, batch size of 64, 10 epochs, with the objective for retrieval, using 3 audio clips during both training and testing.
        \item \textbf{AudioCaps}: Learning rate of 2e-5, batch size of 64, 10 epochs, with the objective for retrieval, using 1 audio clip during both training and testing.
    \end{itemize}
    \item \textbf{Multi-modal (MM)}
    \begin{itemize}
        \item \textbf{MSRVTT}: Learning rate of 2e-5, batch size of 64, 3.6 epochs, with the objective for retrieval, using 8 video frames during training and 16 during testing, with a resolution of 224.
        \item \textbf{YouCook2}: Learning rate of 3e-5, batch size of 64, 30 epochs, with the objective for retrieval, using 8 video frames during training and 16 during testing, with a resolution of 224.
        \item \textbf{VALOR-32K}: Learning rate of 2e-5, batch size of 64, 10 epochs, with the objective for retrieval, using 8 video frames during both training and testing, with a resolution of 224.
        \item \textbf{VATEX}: Learning rate of 2e-5, batch size of 64, 2.5 epochs, with the objective for retrieval, using 8 video frames during training and 16 during testing, with a resolution of 224.
        \item \textbf{DiDeMo}: Learning rate of 2e-5, batch size of 64, 40 epochs, with the objective for retrieval, using 8 video frames during training and 32 during testing, and 2 audio clips during both training and testing, with a resolution of 224.
        \item \textbf{ANET}: Learning rate of 2e-5, batch size of 64, 20 epochs, with the objective for retrieval, using 8 video frames during training and 32 during testing, and 2 audio clips during both training and testing, with a resolution of 224.
    \end{itemize}
\end{itemize}

\textbf{Captioning Tasks (CAP)}
\begin{itemize}
    \item \textbf{Image-Text Modality}
    \begin{itemize}
        \item \textbf{MSCOCO}: Learning rate of 1e-5, batch size of 64, 5 epochs, with the objective for caption, and a resolution of 480.
        \item \textbf{MSCOCO(SCST)}: Learning rate of 2.5e-6, batch size of 64, 2.5 epochs, with the objective for caption, and a resolution of 480.
    \end{itemize}
    \item \textbf{Audio-Text Modality (A-T)}
    \begin{itemize}
        \item \textbf{ClothoV1/V2}: Learning rate of 2e-5, batch size of 64, 10 epochs, with the objective for caption, using 3 audio clips during both training and testing.
        \item \textbf{AudioCaps}: Learning rate of 2e-5, batch size of 64, 10 epochs, with the objective for caption, using 1 audio clip during both training and testing.
    \end{itemize}
    \item \textbf{Multi-modal (MM)}
    \begin{itemize}
        \item \textbf{MSRVTT}: Learning rate of 2e-5, batch size of 128, 10 epochs, with the objective for caption, using 8 video frames during both training and testing, with a resolution of 224.
        \item \textbf{YouCook2}: Learning rate of 3e-5, batch size of 64, 30 epochs, with the objective for caption, using 8 video frames during training and 16 during testing, with a resolution of 224.
        \item \textbf{VALOR-32K}: Learning rate of 1e-5, batch size of 64, 10 epochs, with the objective for caption, using 8 video frames during training and 12 during testing, with a resolution of 224.
    \end{itemize}
\end{itemize}

\textbf{Question Answering Tasks (QA)}
\begin{itemize}
    \item \textbf{Visual-Text Modality (Vis)}
    \begin{itemize}
        \item \textbf{MSVD-QA}: Learning rate of 1e-5, batch size of 64, 10 epochs, with the objective for QA, using 8 video frames during training and 14 during testing, with a resolution of 224.
        \item \textbf{TGIF-FrameQA}: Learning rate of 2e-5, batch size of 64, 10 epochs, with the objective for QA, using 4 video frames during both training and testing, with a resolution of 224.
        \item \textbf{VQAv2}: Learning rate of 2e-5, batch size of 128, 20 epochs, with the objective for QA, and a resolution of 384.
    \end{itemize}
    \item \textbf{Multi-modal (MM)}
    \begin{itemize}
        \item \textbf{MSRVTT-QA}: Learning rate of 2e-5, batch size of 64, 4.5 epochs, with the objective for QA, using 8 video frames and 1 audio clip during both training and testing, with a resolution of 224.
        \item \textbf{MUSIC-AVQA}: Learning rate of 2e-5, batch size of 64, 20 epochs, with the objective for QA, using 8 video frames and 2 audio clips during both training and testing, with a resolution of 224.
        \item \textbf{ANET-QA}: Learning rate of 2e-5, batch size of 64, 10 epochs, with the objective for QA, using 8 video frames during training and 16 during testing, and 2 audio clips during both training and testing, with a resolution of 224.
    \end{itemize}
\end{itemize}

These settings have been optimized to balance efficiency and performance, even though most hyper-parameters are not precisely tuned.

\begin{table*}[t]
    \centering
    \renewcommand\arraystretch{1.0}
    \footnotesize
    \caption{\textbf{Detailed training configurations of \mico for multimodal learning.} Apart from the configurations shown in the table, for image tasks, we use random left-right flipping, random resized crop, color jitter of 0.4, Auto-augment, and no repeated augmentation for every model.}
    \resizebox{0.98\linewidth}{!}{
\begin{tabular}{@{\ }l|cc|cc|cc|cc}
\hline
\multirow{2}{*}{settings} & \multicolumn{2}{c|}{Image} & \multicolumn{2}{c|}{Audio} & \multicolumn{2}{c|}{Video} & \multicolumn{2}{c}{Depth \& Normal Map} \\
\cline{2-9}
& 
ViT-L & 
ViT-g &
ViT-L & 
ViT-g &
ViT-L & 
ViT-g &
ViT-L & 
ViT-g \\
\hline
Input Shape & 
224 & 
224 &
224 & 
224 & 
224 & 
224 &
224 & 
224 \\
batch size & 
4096 &
512 &
4096 &
512 &
4096 &
512 &
4096 &
512 \\
optimizer &
AdamW & 
AdamW &
AdamW &
AdamW &
AdamW &
AdamW & AdamW & AdamW\\
LR      & 
4$\times10^{-3}$ &
5$\times10^{-5}$ &
4$\times10^{-3}$ &
5$\times10^{-5}$ & 
4$\times10^{-3}$ &
5$\times10^{-5}$ & 
4$\times10^{-3}$ &
5$\times10^{-5}$ \\
LR schedule& 
cosine  &
cosine & 
cosine & 
cosine & 
cosine &
cosine & cosine & cosine\\
weight decay     &
0.05  &
1$\times10^{-8}$ & 
0.05  &
1$\times10^{-8}$ & 
0.05  &
1$\times10^{-8}$ & 
0.05  &
1$\times10^{-8}$ \\ 
warmup epochs & 
5 &
0 &
5 &
0 &
5 &
0 &
5 &
0 \\
epochs & 
90 &
30 &
90 &
30 &
90 &
20 &
90 &
20 \\
\hline

mixup alpha  & 
0.8 & 
0.0 &
0.8 & 
0.0 &
0.8 & 
0.0 &
0.8 & 
0.0 \\
cutmix alpha &
1.0 & 
0.0 &
1.0 & 
0.0 &
1.0 & 
0.0 &
1.0 & 
0.0 \\
erasing prob. &
0.25    &
0.25   &
0.25 &
0.25 &
0.25    &
0.25   &
0.25 &
0.25 \\
\hline
dropout rate & 
0.1 & 
0.2 & 
0.1 & 
0.2 &
0.1 &
0.3 & 
0.2 & 
0.3 \\
\hline
\end{tabular}
}
\label{tab:hyper}
\end{table*}

\begin{center}
\vskip -0.15in
\begin{algorithm}[t]
    \caption{Multimodal Context Pretraining Algorithm, PyTorch-like}
    \label{alg:code}
    \definecolor{codeblue}{rgb}{0.25,0.5,0.5}
    \definecolor{codekw}{rgb}{0.85, 0.18, 0.50}
    \lstset{
      language=Python,
      basicstyle=\small\ttfamily,
      columns=fullflexible,
      breaklines=true,
      postbreak=\mbox{},
      keywordstyle=\color{blue},
      commentstyle=\color{green},
      stringstyle=\color{red},
      stepnumber=1,
      numbersep=10pt,
      showspaces=false,
      showstringspaces=false,
      tabsize=3,
      captionpos=b,
      frame=single
    }
    
    \begin{lstlisting}[language=python]
def train(video_pixels=None, image_pixels=None, depth_pixels=None, audio_spectrograms=None):
    # Get Mixed Data
    modal_inputs = [video_pixels, image_pixels, depth_pixels, audio_spectrograms]
    modal_captions = [video_captions, image_captions, depth_captions, audio_captions]

    # Extract Features
    modal_feats = [self.encoder(modal) for modal in modal_inputs if modal is not None]
    multimodal_feats = torch.cat(modal_feats)
    concatenated_captions = ''.join(modal_captions)
    text_feats = self.text_encoder(concatenated_captions)

    # Losses
    contra_loss = Contrasive_Loss(multimodal_feats, text_feats)
    matching_loss = Matching_Loss(modal_captions, multimodal_feats)
    gen_loss = Generation_Loss(modal_captions.mask(0.6), multimodal_feats)

    # Total Loss
    loss = contra_loss + matching_loss + gen_loss

return loss
\end{lstlisting}
\end{algorithm}
\vskip -0.2in
\end{center}

For evaluation purposes, we employ different strategies tailored to specific tasks:

1. Retrieval Tasks: All candidates are initially ranked using Omni-modal Contrastive Loss. Following this, the Top-50 candidates undergo a reranking process through the Omni-modal Matching Process.

2. Captioning Tasks: Beam search with a beam size of 3 is utilized to generate captions, ensuring a comprehensive exploration of possible outputs.

3. Question Answering (QA) Tasks: These are treated as open-ended generative problems. Questions are used as prefixes, and answers are generated without any constraints, allowing for flexible and contextually appropriate responses.

For comparisons with state-of-the-art (SOTA) models and ablation studies, we use the following evaluation metrics: 1) Retrieval Tasks: Recall@1. 2) Captioning Tasks: CIDEr. 3) QA Tasks: Accuracy (Acc)
These metrics provide a comprehensive assessment of the model's performance across different types of tasks.

\section{Datasets and Metrics}~\label{sec:app:data}
\textbf{Dataset Split}
To split the mix of datasets into subsets of 1M, 10M, 110M, and 334M video clips while preserving its diversity and quality, we employed a proportional stratified sampling method. Initially, the dataset, which spans over 15 categories (including music, gaming, education, entertainment, and animals) and includes vision, audio, depth, normal maps, and text modalities, was organized and labeled. Stratified random sampling was then used to ensure each subset accurately reflected the distribution of categories and modalities present in the full dataset. This method involved selecting samples proportionally from each category to maintain representative distributions. The vision and audio captions were also kept proportional in length and quantity, ensuring that each subset retained the comprehensive characteristics of the original dataset.

\begin{center}
\vskip -0.15in
\begin{algorithm}[t]
    \caption{Dataset Split Algorithm}
    \label{alg:split}
    \definecolor{codeblue}{rgb}{0.25,0.5,0.5}
    \definecolor{codekw}{rgb}{0.85, 0.18, 0.50}
    \lstset{
      language=Python,
      basicstyle=\small\ttfamily,
      columns=fullflexible,
      breaklines=true,
      postbreak=\mbox{},
      keywordstyle=\color{blue},
      commentstyle=\color{green},
      stringstyle=\color{red},
      stepnumber=1,
      numbersep=10pt,
      showspaces=false,
      showstringspaces=false,
      tabsize=3,
      captionpos=b,
      frame=single
    }
    
    \begin{lstlisting}[language=python]
import pandas as pd
from sklearn.model_selection import train_test_split

# Assume `data` is a DataFrame containing the full dataset with columns ['category', 'vision_caption', 'audio_caption', 'depth', 'normal', 'subtitle']
# Adding an 'index' column to keep track of the original indices
data['index'] = data.index

# Define the sizes of each subset
subset_sizes = [1e6, 1e7, 1.1e7, 3.34e7]

# Function to create stratified samples
def create_subset(data, size):
    subset, _ = train_test_split(data, train_size=size, stratify=data['category'], random_state=42)
    return subset

# Creating subsets
subset_1M = create_subset(data, 1e6)
subset_10M = create_subset(data, 1e7)
subset_110M = create_subset(data, 1.1e7)
subset_334M = create_subset(data, 3.34e7)

# Reset index for each subset 
subset_1M.reset_index(drop=True, inplace=True)
subset_10M.reset_index(drop=True, inplace=True)
subset_110M.reset_index(drop=True, inplace=True)
subset_334M.reset_index(drop=True, inplace=True)

\end{lstlisting}
\end{algorithm}
\vskip -0.2in
\end{center}

\subsection{Single-modality Evaluation Details}
{\color{TextDark}{\textbf{Text}}}. The MMLU (Massive Multitask Language Understanding) benchmark is designed to evaluate the multitask accuracy of language models across 57 diverse tasks, including subjects like mathematics, history, and biology. It assesses models' abilities to generalize and apply knowledge in various domains, providing a comprehensive measure of text understanding and reasoning skills.

{\color{ImageDark}{\textbf{Image}}}. We conduct experiments on ImageNet-1K~\cite{deng2009imagenet}, a dataset comprising approximately 1.3 million images across 1,000 categories. In line with common practices~\cite{wang2021pyramid,liu2021swin,liu2022convnet,ding2023unireplknet}, base-scale models are trained for 300 epochs. Large-scale models undergo pre-training on ImageNet-22K, which includes 14.2 million images, for 90 epochs, followed by fine-tuning on ImageNet-1K for an additional 20 epochs.

{\color{ThermalDark}\textbf{Thermal and Hyperspectral data understanding}}. We conduct experiments on infrared image recognition using the RegDB dataset, X-ray scan analysis with the Chest X-Ray dataset~\cite{rahman2020reliable}, and hyperspectral data recognition using the Indian Pine dataset\footnote{\url{https://github.com/danfenghong/IEEE_TGRS_SpectralFormer/blob/main/data/IndianPine.mat}}.

{\color{DepthDark}{\textbf{Depth}}}. 
The NYU Depth Dataset (NYU-D) comprises RGB and depth image pairs captured from indoor scenes. It includes 1,449 densely labeled pairs for training and testing, along with over 400,000 unlabeled frames.

{\color{AudioDark}{\textbf{Audio}}}. For audio recognition, Audioset-2M dataset comprises over 2 million human-labeled 10-second audio clips drawn from YouTube videos. It covers a wide range of 527 sound event classes, providing a comprehensive resource for training and evaluating audio event detection and classification models.

{\color{VideoDark}{\textbf{Video}}}. The Kinetics-700 dataset contains 700,000 video clips covering 700 human action classes, used for action recognition tasks. The MSR-VTT dataset includes 10,000 video clips paired with multiple textual descriptions, supporting video captioning, retrieval, and content understanding research.

{\color{TimeDark}{\textbf{Time-series}}}. Global Weather Forecasting~\cite{wu2023interpretable} includes global, regional, and Olympics data from NCEI and CMA, comprising hourly weather measurements from thousands of stations. Evaluation involved splitting data into training, validation, and test sets (7:1:2) using MSE and MAE metrics.

{\color{GraphDark}{\textbf{Graph}}}. PCQM4M-LSC dataset is a large-scale collection of 4.4 million organic molecules, each with up to 23 heavy atoms and associated quantum-mechanical properties. Aimed at predicting molecular properties through machine learning, this dataset is highly relevant for applications in drug discovery and material science.

{\color{TabularDark}{\textbf{Tabular}}}. The fraud dataset comprises transaction records, including features like transaction amount, location, time, and user information. It is designed for machine learning models to detect fraudulent activities. This dataset is crucial for developing and testing algorithms to enhance security in financial systems and reduce economic losses due to fraud. 

{\color{IMUDark}{\textbf{IMU}}}. The Ego4D dataset includes inertial measurement unit (IMU) data captured from wearable devices, providing detailed motion and orientation information. This dataset supports research in human activity recognition, augmented reality, and robotics, offering comprehensive insights into human movements and interactions with the environment.

\subsection{Cross-modality Evaluation Details}

We evaluated \mico across several well-known downstream datasets, including MSRVTT, VATEX, YouCook2, VALOR-32K, MSVD, DiDeMo, ActivityNet Caption, TGIF, MUSIC-AVQA, Clotho, AudioCaps, MSCOCO, Flickr30K, and VQAv2. The specific train/validation/test splits for these benchmarks are detailed below:

\section*{Retrieval Tasks}

\subsection*{Audio-Text Modality (A-T)}
\begin{itemize}
    \item \textbf{ClothoV1}~\cite{drossos2020clotho}: This dataset includes 2,893 audio clips for training and 1,045 for validation. The corresponding captions number 14,465 for training and 5,225 for validation.
    \item \textbf{ClothoV2}~\cite{drossos2020clotho}: Contains 3,839 audio clips for training and 1,045 for validation, with 19,195 captions for training and 5,225 for validation.
    \item \textbf{AudioCaps}~\cite{kim2019audiocaps}: Comprises 49,291 audio clips for training, 428 for validation, and 816 for testing, along with 49,291 captions for training, 2,140 for validation, and 4,080 for testing.
\end{itemize}

\subsection*{Video-Text Modality (V-T)}
\begin{itemize}
    \item \textbf{MSRVTT}~\cite{xu2016msr}: Comprises 10K video clips and 200K captions, spanning diverse topics such as human activities, sports, and natural landscapes. We evaluate text-to-video retrieval, video captioning, and video QA using this dataset. Contains 9,000 videos for training and 1,000 for testing, with 180,000 captions for training and 1,000 for testing.
    \item \textbf{YouCook2}~\cite{zhou2018towards}: Comprises 14K video clips extracted from 2K instructional cooking videos on YouTube. Each video features multiple actions performed by chefs, along with corresponding textual descriptions and temporal annotations. Includes 10,337 videos for training and 3,492 for validation, with matching captions.
    \item \textbf{VALOR-32K}~\cite{chen2023valor}: An audiovisual video-language benchmark containing 32K 10-second video clips sourced from AudioSet~\cite{gemmeke2017audio}. Each clip includes annotations with captions that describe both the visual and audio content. Consists of 25,000 videos for training, 3,500 for validation, and 3,500 for testing, each with corresponding captions.
    \item \textbf{DiDeMo}~\cite{anne2017localizing}: Comprises 10K long-form videos sourced from Flickr, with each video annotated with four short sentences in temporal order. For this benchmark, we concatenate these short sentences and evaluate 'paragraph-to-video' retrieval, using the official split. Features 8,394 videos for training, 1,065 for validation, and 1,003 for testing, along with their captions.
    \item \textbf{ActivityNet (ANET)}~\cite{krishna2017dense}: Includes 20K long-form videos (average length of 180 seconds) from YouTube, accompanied by 100K captions. We evaluate text-to-video retrieval and video QA on this dataset. Comprises 10,009 videos for training and 4,917 for testing, with corresponding captions.
    \item \textbf{LSMDC}~\cite{rohrbach2017movie}: Contains 101,046 videos for training, 7,408 for validation, and 1,000 for testing, with corresponding captions.
\end{itemize}

\section*{Captioning Tasks}

\subsection*{Audio-Text Modality (A-T)}
\begin{itemize}
    \item \textbf{ClothoV1}~\cite{drossos2020clotho}: This dataset includes 2,893 audio clips for training and 1,045 for validation. The corresponding captions number 14,465 for training and 5,225 for validation.
    \item \textbf{ClothoV2}~\cite{drossos2020clotho}: Contains 3,839 audio clips for training and 1,045 for validation, with 19,195 captions for training and 5,225 for validation.
    \item \textbf{AudioCaps}~\cite{kim2019audiocaps}: Comprises 49,838 audio clips for training, 495 for validation, and 975 for testing, along with 49,438 captions for training, 2,475 for validation, and 4,875 for testing.
\end{itemize}

\subsection*{Video-Text Modality (V-T)}
\begin{itemize}
    \item \textbf{MSRVTT}~\cite{xu2016msr}: Contains 6,513 videos for training, 497 for validation, and 2,990 for testing, with 130,260 captions for training, 9,940 for validation, and 59,800 for testing.
    \item \textbf{YouCook2}~\cite{zhou2018towards}: Includes 10,337 videos for training and 3,492 for validation, with matching captions.
    \item \textbf{VALOR-32K}~\cite{chen2023valor}: Consists of 25,000 videos for training, 3,500 for validation, and 3,500 for testing, each with corresponding captions.
    \item \textbf{VATEX}~\cite{wang2019vatex}: Consists of 41,250 video clips sourced from the Kinetics-600 dataset~\cite{kay2017kinetics}, accompanied by 825,000 sentence-level descriptions. Contains 25,991 videos for training, 3,000 for validation, and 6,000 for testing, with 259,910 captions for training, 30,000 for validation, and 60,000 for testing.
\end{itemize}

\section*{Question Answering (QA) Tasks}

\subsection*{Video-Text Modality (V-T)}
\begin{itemize}
    \item \textbf{MSRVTT-QA}~\cite{xu2017video}: Contains 6,513 videos for training, 497 for validation, and 2,990 for testing, with 158,581 QA pairs for training, 12,278 for validation, and 72,821 for testing.
    \item \textbf{MUSIC-AVQA}~\cite{li2022learning}: An audiovisual video QA benchmark containing over 45K Q-A pairs, covering 33 different question templates across various modalities and question types. Includes 9,277 videos for training, 3,815 for validation, and 6,399 for testing, with 32,087 QA pairs for training, 4,595 for validation, and 9,185 for testing.
    \item \textbf{ANET-QA}~\cite{yu2019activitynet}: Comprises 3,200 videos for training, 1,800 for validation, and 800 for testing, with 32,000 QA pairs for training, 18,000 for validation, and 8,000 for testing.
\end{itemize}

\section*{Image-Based Tasks}

\begin{itemize}
    \item \textbf{MSCOCO}~\cite{lin2014microsoft}: Comprises 123K images, each paired with 5 annotated captions. We evaluate text-to-image retrieval and image captioning on this dataset.
    \item \textbf{Flickr30K}~\cite{plummer2015flickr30k}: Contains 31K images, each paired with five descriptive captions. This dataset is widely used for evaluating image captioning and text-to-image retrieval tasks.
\end{itemize}

\section*{Visual Question Answering}

\begin{itemize}
    \item \textbf{VQAv2}~\cite{goyal2017making}: A large-scale Visual Question Answering dataset comprising over 265K images and 1.1M questions, designed to improve the balance of answer types per question. This dataset is used to evaluate models' abilities to understand and reason about visual content by providing accurate answers to questions based on the images.
\end{itemize}


\end{document}